\documentclass{article}
\PassOptionsToPackage{hyphens}{url}
\usepackage{iclr2027_conference,times}
\usepackage{amsmath,amsfonts,bm}

\def\eqref#1{equation~\ref{#1}}
\def\1{\bm{1}}

\DeclareMathAlphabet{\mathsfit}{\encodingdefault}{\sfdefault}{m}{sl}
\SetMathAlphabet{\mathsfit}{bold}{\encodingdefault}{\sfdefault}{bx}{n}

\usepackage{hyperref}
\usepackage{url}
\usepackage{booktabs}
\usepackage{multirow}
\usepackage{amsmath, amssymb}
\usepackage{graphicx}
\usepackage{xcolor}
\usepackage{xspace}
\usepackage{tikz}
\usetikzlibrary{positioning,arrows.meta,calc,backgrounds,fit,shapes.geometric}
\usepackage{subcaption}
\usepackage{wrapfig}
\usepackage{enumitem}
\usepackage[capitalize,nameinlink]{cleveref}

\definecolor{slate}{HTML}{CBD5E1}\definecolor{teal}{HTML}{99D6C9}\definecolor{amber}{HTML}{F6D28B}
\definecolor{indigo}{HTML}{B7C4F2}\definecolor{apricot}{HTML}{F8C9A0}\definecolor{rose}{HTML}{F4B6BC}\definecolor{lilac}{HTML}{D9C8F0}\definecolor{ink}{HTML}{475569}

\newcommand{\bench}{SLIM\xspace}
\newcommand{\lewm}{LeWM\xspace}
\newcommand{\ours}{LeWM+I\xspace}
\newcommand{\pp}{\,pp\xspace}
\newcommand{\sgop}{\mathrm{sg}}
\newcommand{\Lpred}{\mathcal{L}_\text{pred}}
\newcommand{\Lenc}{\mathcal{L}_\text{enc-IDM}}
\newcommand{\Lpid}{\mathcal{L}_\text{pred-IDM}}
\title{Keeping JEPA World Models Plannable\\When Little of the Frame Moves}

\author{Florian Strohm$^{1}$, Patrick Wagner$^{1}$, Jannik Schwab$^{1}$ \& Marco Huber$^{1,2}$ \\
$^{1}$Fraunhofer IPA, Stuttgart, Germany \qquad $^{2}$University of Stuttgart, Germany \\
\texttt{florian.strohm@ipa.fraunhofer.de}}

\iclrfinalcopy
\begin{document}

\maketitle
\lhead{Preprint}

\begin{abstract}
Specifying a goal in language rather than as a goal frame is a natural interface for planning with a latent world model, but testing it needs scenes in which language must discriminate between several objects. We build \bench, a pushing benchmark with several small objects and paired visual and language goals on identical scenes. On \bench a \lewm world model that solves PushT succeeds on under $1\%$ of trials, although a scripted controller with simulator state solves every tier. Probes locate the failure in the encoder: its latent is nearly action-insensitive, neither pusher nor object positions can be decoded from it, and rollouts are no better than copying the current latent forward. One inverse-dynamics auxiliary loss, applied to encoder latents and to predicted latents through a shared head discarded at test time, restores every probe and raises success from $0.003$ to $0.35$ ($0.16$ on the hard pushing tier, where a goal-agnostic policy scores zero), and improves PushT at twice the trained horizon. Controls attribute the repair to the gradient into the encoder, and a response sweep shows that the vanilla model plans once enough of the frame responds to actions. A cheap action-sensitivity probe, computable without environment access, acts as an empirical necessary condition: all configurations below its threshold failed to plan. On the repaired latent, a small language-goal head plans from sentences without retraining the world model: it reaches $0.84$ on navigation (visual-goal oracle $1.00$), follows the named zone when it is swapped with a decoy, and degrades gracefully to unseen nouns. A single goal sentence rarely completes a push, but given the push as a sequence of stage sentences the head raises success on the medium and hard pushing tiers from $0.04$ to $0.25$, on par with the goal-frame oracle, also when the switch between stages is read from the latent alone.
\end{abstract}

\section{Introduction}
\label{sec:intro}

Visual planning agents in the wild rarely have a pixel-precise goal frame to plan toward; a \emph{language description} of what to achieve is a far more natural interface. We set out to build such a planner on a JEPA-style latent world model \citep{lecun2022path,assran2023ijepa}, a sample-efficient recipe for visual control \citep{maes2026leworldmodel,sobal2025pldm,zhou2024dinowm} that predicts latents rather than pixels \citep{ha2018worldmodels,hafner2023dreamerv3}.

Two prerequisites must hold. (1)~The benchmark must be visually diverse enough that language can \emph{discriminate}; PushT \citep{chi2023diffusionpolicy}, with a single object and a single goal region, is not rich enough to support linguistically diverse instructions. (2)~The latent must be \emph{plannable}: the encoder must keep action-relevant variance so a sampling-based planner has a descent direction. We built \bench (small-object, language-instructed, multi-object pushing) for the first prerequisite and found that the second fails when little of the frame responds to actions.

\begin{figure}[t]
\centering
\includegraphics[width=0.42\textwidth]{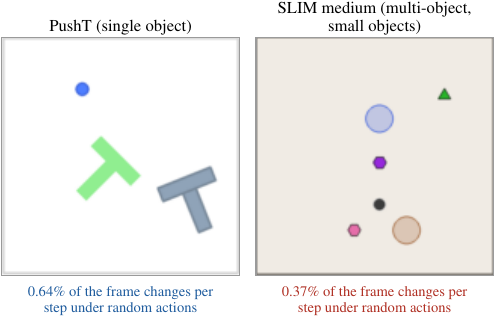}
\caption{\textbf{Why JEPA world models collapse on \bench.} Under the same random actions, $0.64\%$ of the frame changes per predictor step on PushT (left), where the large T-block moves whenever it is touched, and $0.37\%$ on \bench (right; the same scene as in \cref{fig:scenes}), where the pusher rarely hits one of several objects. When so little of the frame responds to actions, the prediction objective can be minimised by an action-insensitive encoder (\cref{sec:mechanism}).}
\label{fig:teaser}
\end{figure}

A vanilla JEPA world model (\lewm, \citealt{maes2026leworldmodel}) that reaches $98\%$ success on PushT succeeds on 3 of 1200 \bench trials. Probes locate the failure in the encoder: the latent is nearly action-insensitive ($1500\times$ below the repaired model), neither the pusher nor the objects can be decoded from it, the predictor's rollout error exceeds the true latent motion tenfold, and the planner's cost landscape is flat. We trace this to how little of the frame responds to actions (\cref{fig:teaser}): the pusher rarely touches one of several objects, which then move $24$--$50\times$ less than PushT's T-block under sampled actions, so a representation that ignores all bodies costs the prediction loss little, and the isotropic regulariser (SIGReg) prevents global collapse without choosing \emph{which} variance survives. Consistent with this account, doubling the object radius changes nothing, whereas making more of the frame respond to actions lifts the vanilla model from $0.005$ to $0.530$.

We repair the latent with a single \textbf{inverse-dynamics auxiliary loss (I)}: one shared inverse-dynamics model (IDM) head predicts the action from consecutive encoder latents and from (current, predicted) latent pairs, and is discarded at test time. \ours lifts \bench success from $0.003$ to $0.348\pm0.016$ (mean $\pm$ SE over three training seeds) and PushT success at twice the trained horizon by $+30$\pp. Controls show that \emph{the gradient into the encoder carries the effect}: the encoder-side term alone recovers most of the gain, the predictor-side term alone none, and together they are best (\cref{sec:disentangle}).

With a plannable latent we return to language: a small head maps the current latent and a frozen sentence embedding of the goal to a goal latent for the planner's cost, leaving the world model untouched. Attending over the encoder's patch tokens, it reaches $0.84$ on navigation against the visual-goal oracle's $1.00$, ends closer to the named zone than to a swapped decoy in all 300 decoy episodes, and keeps $84\%$ of its success with unseen nouns; from one goal sentence it reaches about a fifth of the oracle on the pushing tiers. Given the push as a sequence of stage sentences instead, it pushes as often as the goal-frame oracle ($0.25$ on medium and hard, oracle $0.27$ and $0.16$), also when the switch between stages is read from the latent alone. An oracle given the stage frames ($0.62$ and $0.55$) shows that decomposing the push helps the planner itself and marks the remaining headroom; unseen nouns still remove most of the staged gain.

\paragraph{Contributions.}
\begin{itemize}[leftmargin=1.5em,itemsep=1pt,topsep=2pt]
\item \bench, a multi-object 2D pushing benchmark with four difficulty tiers and paired visual and language goals on identical scenes, which exposes the regime in which little of the frame responds to actions, at PushT-level compute (\cref{sec:bench}).
\item A mechanistic account of \emph{when and why} \lewm collapses (one model-free and four model-internal probes, and a response-to-actions sweep), an inverse-dynamics auxiliary loss on encoder and predicted latents that repairs it, and controls (encoder-vs-predictor disentangle, shuffled-action null, frozen-encoder cap) that pin the repair to encoder shaping (\cref{sec:method,sec:exp}).
\item A language-goal head that plugs into the planner without retraining the world model, approaches the visual-goal oracle on navigation, grounds the named zone in a decoy test, degrades gracefully to unseen nouns and, given the push as stage sentences, pushes as often as the goal-frame oracle (\cref{sec:nl}).
\end{itemize}

\section{Related work}
\label{sec:related}

\paragraph{JEPA world models for control.} Joint-embedding predictive architectures \citep{lecun2022path,assran2023ijepa,bardes2024vjepa,assran2025vjepa2} predict embeddings instead of pixels and regularise against collapse \citep{bardes2022vicreg,balestriero2025lejepa}. For control, DINO-WM \citep{zhou2024dinowm} trains a predictor over a frozen DINOv2 encoder \citep{oquab2024dinov2}, PLDM \citep{sobal2025pldm} plans in a VICReg-regularised JEPA latent on state-based navigation, and \lewm \citep{maes2026leworldmodel} trains a ViT encoder end-to-end with the SIGReg regulariser of LeJEPA \citep{balestriero2025lejepa} \citep[see also][]{terver2025jepawm}. Inverse models as auxiliary objectives go back to learning to poke \citep{agrawal2016poke} and curiosity \citep{pathak2017icm}; the controllable-latent-state literature uses them to keep what the agent controls and discard exogenous noise, and shows that one-step inverse models can miss parts of the controllable state \citep{lamb2022acstate,efroni2022exbmdp,islam2023acro,wang2022denoisedmdp,levine2024multistep,yeom2026actionrelevant}. Our one-step IDM makes the pusher decodable; that it also restores the objects, which the action does not determine, is an empirical finding, and the weak pushing tiers are consistent with the limits of one-step inverse models.

\paragraph{Concurrent work.} Concurrently with and independently of this work, several 2026 preprints add an inverse-dynamics objective on consecutive encoder latents to \lewm-style models. DA-LeWM \citep{wang2026dalewm} adds an MSE IDM head at weight 0.1 with a goal-conditioned action head and studies convergence speed and cost geometry on PushT, Reacher, Cube and TwoRoom, where \lewm already plans. ATM \citep{chen2026atm} also tries a head on the predicted latent, but as a substitute for the encoder term rather than a shared head with stop-gradient, and finds it a shortcut. SMWM \citep{ivashkov2026smwm}, Delta-JEPA \citep{zhang2026deltajepa}, AC-MTM \citep{boylan2026acmtm} and \citet{liu2026physgroundedjepa} \emph{replace} SIGReg by an MSE, latent-difference or contrastive IDM and see collapse only once the regulariser is removed; AC-MTM reports IDM \emph{failing} where actions barely move a variable (PushT block orientation), whereas in our setting the whole latent collapses and IDM repairs it. None of these works studies scenes in which little of the frame responds to actions, reports a shuffled-action null, a frozen-encoder control or an encoder-vs-predictor disentangle, or plans toward language goals. ACPC \citep{an2026acpc} and ActSWM \citep{gan2026actswm} target action sensitivity as a post-hoc screen and as a rollout constraint, respectively. Other latent-shaping objectives \citep{wang2026straightening,bai2026tdjepa,li2026rcaux,zeng2026phylatent,toso2026bisimjepa} and object-centric latents \citep{nam2026causaljepa,daniel2026lpwm,spieler2026slotmpc,zhang2026dila} are complementary.

\paragraph{Language goals for latent planning.} Language has been used as a learned reward for visual model-predictive control \citep{nair2021lorel}, as a goal space shared with goal images for policies \citep{lynch2021langlfp,mees2022hulc}, and to generate subgoal images \citep{black2024susie}. DynaLang \citep{lin2024dynalang} predicts language tokens inside a Dreamer world model. Closest to our head is LAGO \citep{barbeau2026lago}, which maps a frozen sentence embedding to latent (sub)goals for CEM by co-training the predictor over a frozen DINOv2 encoder. Our head is a separate 1--5M-parameter module trained on a frozen world model and is tested under paraphrase, which LAGO, GWM \citep{li2026gwm} and SAGE \citep{cheng2026sage} do not report. PiJEPA \citep{chahe2026pijepa} uses language as a policy prior for model-predictive path-integral (MPPI) control, not as the planner's goal. Language-model planners decompose an instruction into steps for a low-level controller \citep{huang2022zeroshot,ahn2022saycan}; our staged instructions (\cref{sec:staged}) hand such steps to a latent planner as subgoal latents.

\paragraph{Benchmarks.} PushT \citep{chi2023diffusionpolicy}, BlockPush and Sawyer variants are single-object; OGBench's visual scene tasks \citep{park2024ogbench} are multi-object at $64\times64$, with language labels added by LAGO. Language-Table \citep{lynch2023interactive} and CALVIN \citep{mees2022calvin} pair multi-object tabletop manipulation with far richer language; what \bench adds is a controlled regime in which little of the frame responds to actions, at PushT-level compute.

\section{The \bench benchmark}
\label{sec:bench}

\begin{figure}[t]
\centering
\includegraphics[width=0.8\textwidth]{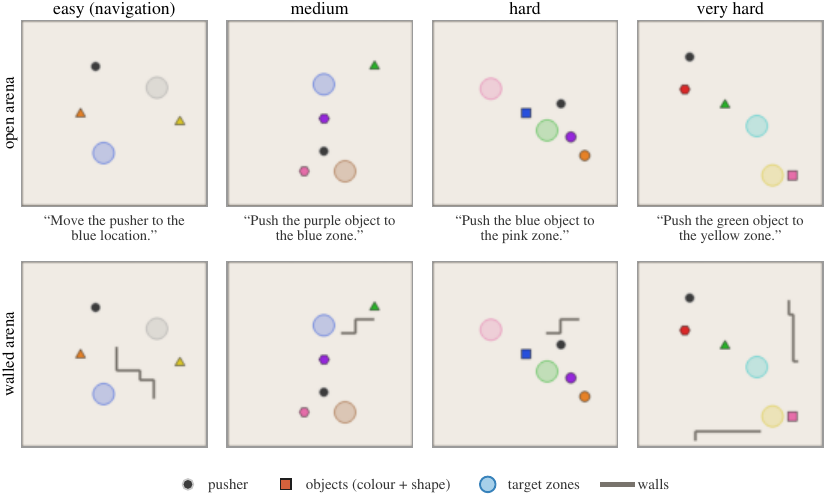}
\caption{\bench scenes for the four tiers (columns) in the open arena (top) and the identical configuration with procedurally generated wall chains added (bottom). The templated language goal of each scene is quoted under the top row; the easy tier asks only to move the pusher. Tiers differ in the geometry of pusher, target object and zone, not in object appearance.}
\label{fig:scenes}
\end{figure}

\paragraph{Design.} Standard world-model benchmarks sit in a regime where sampled actions change a sizeable part of the frame, so a world model can look strong without ever being tested on scenes where little moves. \bench is a controlled testbed for that regime at PushT-level compute: several small objects per scene with a controlled distractor count, tiers from solved to unsolved, simulator constants that can be tuned to vary how much of the frame responds to actions (\cref{sec:mechanism}), a scripted controller as a ceiling, and paired visual and language goals on identical scenes.

The arena is a top-down 2D scene rendered at $224\times224$ and simulated with pymunk. The pusher is a 14\,px-radius disc driven by a 2D target velocity $(\mathrm{d}x,\mathrm{d}y)\in[-1,1]^2$ through a PD controller; objects are 15\,px-radius bodies drawn from four shapes and a colour palette, damped so they stop shortly after contact. Every scene has two coloured zones, one of which is the target. Four tiers differ in geometry: \emph{easy} is a navigation tier (pusher to target zone; 0--2 distractor objects, nothing to push); \emph{medium} places pusher, target object and zone collinearly; \emph{hard} adds $\sim25$\,px lateral offsets so the push must be steered; \emph{very hard} puts pusher, object and zone far apart along a diagonal. Push tiers contain 1--3 objects of distinct colour, one of them the target. The tier's template (pusher, target object, target zone) is placed at a random position in the arena, distractor objects and zones are drawn from the same position distribution as the target object and zone, and the scene is mapped through one of the eight symmetries of the square arena (quarter turns about the centre, optionally mirrored). This varies position and direction of every scene while keeping the template distances, and hence each tier's difficulty, fixed. A policy that ignores the goal and moves in one direction for a tuned number of steps solves $0.12$ of easy and $0.20$ of medium scenes (the symmetries map an axis-aligned template onto four directions, so one direction is aligned with about a quarter of the scenes) and $0$ on hard and very hard (\cref{tab:headline}). Procedural wall chains are off in the headline results (\cref{app:walls}). Success requires, at the end of a 20-control-step episode, the pusher within 30\,px of the target zone (easy) or the target object within 40\,px of it (push tiers). Each scene comes with a templated natural-language goal (``push the purple object to the blue zone'') and with a rendered goal frame, so visual-goal and language-goal planning are evaluated on identical scenes.

For training data we collect 10\,000 episodes of 200 steps with a hand-coded primitive policy that mixes six manoeuvres (orbit, nudge, firm push, glancing hit, direction change, inter-object push) and never sees the goal.

\paragraph{Little of the frame responds to actions.} Before any model is trained, a model-free probe already separates \bench from PushT. Under a shared random-action distribution $\mathcal{N}(0,\sigma^2I)$ clipped to $[-1,1]^2$ at $\sigma{=}0.5$, $0.37\%$ of the $224\times224$ frame changes per predictor step on \bench against $0.64\%$ on PushT (16 start states $\times$ 20 sequences of 25 steps; all bodies counted). The objects account for most of the gap: PushT's T-block is displaced by 27\,px in 25 steps (root-mean-square, 512-px simulator frame), a \bench object by 1.1\,px in a training scene (three to six objects) and 0.54\,px in a single-object scene, because a 14\,px pusher rarely hits 15\,px bodies. The fraction of static-object training transitions does not separate the two benchmarks (\cref{app:traindist}); how much of the frame a moving body changes does.

\section{An inverse-dynamics auxiliary loss for JEPA world models}
\label{sec:method}

\paragraph{Background: \lewm.} A latent JEPA world model \citep{maes2026leworldmodel} consists of an encoder $f_\theta$ with $z_t=f_\theta(x_t)$, an action encoder $g_\phi$, and an autoregressive transformer predictor $p_\psi$ that maps a context window of $H$ latents and actions to the next latent, $\hat z_{t+1}=p_\psi(z_{t-H+1:t},g_\phi(a_{t-H+1:t}))$. Training minimises the latent prediction error plus SIGReg, a regulariser toward an isotropic Gaussian that prevents representational collapse (plain stays collapsed at a third or three times its default weight, \cref{app:alpha}),
\begin{equation}
\mathcal{L}_\text{vanilla}=\underbrace{\tfrac1H\textstyle\sum_{h=1}^{H}\lVert\hat z_{t-H+h+1}-z_{t-H+h+1}\rVert^2}_{\Lpred}+\lambda_\text{sigreg}\,\mathcal{L}_\text{sigreg}(z).
\end{equation}
At test time a cross-entropy-method (CEM) planner rolls sampled action sequences $a_{t:t+H_a-1}$ out in latent space and minimises the path cost $\tfrac{1}{H_a}\sum_{k=1}^{H_a}\lVert\hat z_{t+k}-z_\text{goal}\rVert^2$ against the encoded goal frame (the mean over the rollout; a terminal-only cost is far worse, \cref{app:cadence}).

This suffices on PushT, where one large block moves whenever the pusher touches it. On \bench the movable bodies contribute little to the frame-to-frame change that $\Lpred$ measures; we hypothesise that the encoder then drops the state of pusher and objects alike, so the predictor copies the current latent forward and the cost surface goes flat (\cref{sec:mechanism}).

% TikZ architecture diagram for the method section.
% \input{} from sec/4_method.tex; tikz libraries are declared in main.tex
% (positioning, arrows.meta, calc, backgrounds, fit, shapes.geometric).
\begin{figure}[t]
\centering
\resizebox{0.93\textwidth}{!}{%
\begin{tikzpicture}[
  font=\footnotesize\sffamily,
  >={Latex[length=1.6mm,width=1.4mm]},
  box/.style   ={draw, line width=0.4pt, rounded corners=2.5pt, align=center, inner sep=3pt,
                 minimum height=5.5mm},
  img/.style   ={box, fill=slate!35, draw=slate!150},
  enc/.style   ={box, fill=teal!35, draw=teal!150, minimum width=8mm},
  act/.style   ={box, fill=amber!45, draw=amber!160},
  pred/.style  ={box, fill=indigo!30, draw=indigo!140, minimum width=12mm},
  lat/.style   ={box, fill=white, draw=slate!150, minimum width=8mm},
  idm/.style   ={box, fill=apricot!60, draw=apricot!185, line width=0.7pt},
  loss/.style  ={box, fill=rose!40, draw=rose!150},
  flow/.style  ={->, line width=0.5pt, rounded corners=3pt, draw=ink!85},
  encf/.style  ={flow, draw=teal!165},
  predf/.style ={flow, draw=indigo!160},
  gradf/.style ={flow, dashed, dash pattern=on 1.6pt off 1.3pt, draw=rose!175},
  tag/.style   ={inner sep=1pt, font=\scriptsize}]

% ---------------- training panel ----------------
\node[act] (at)  at (0.55, 2.95) {$a_t$};
\node[enc] (gphi)at (2.10, 2.95) {$g_\phi$};
\node[img] (xt)  at (0.55, 2.20) {$x_t$};
\node[enc] (f1)  at (2.10, 2.20) {$f_\theta$};
\node[lat] (zt)  at (3.60, 2.20) {$z_t$};
\node[img] (xt1) at (0.55, 1.22) {$x_{t+1}$};
\node[enc] (f2)  at (2.10, 1.22) {$f_\theta$};
\node[lat] (zt1) at (4.55, 1.22) {$z_{t+1}$};
\node[pred](p)   at (6.05, 2.58) {$p_\psi$};
\node[lat] (zh)  at (7.75, 2.58) {$\hat z_{t+1}$};
\node[loss](lp)  at (7.75, 1.22) {$\Lpred$};

\draw[flow] (xt)  -- (f1);
\draw[encf] (f1)  -- (zt);
\draw[flow] (xt1) -- (f2);
\draw[encf] (f2)  -- (zt1);
\draw[flow] (at)  -- (gphi);
\draw[encf] (zt.east) -- ++(0.3,0) |- ([yshift=-1.6mm]p.west);
\draw[flow] (gphi.east) -- ++(0.3,0) |- ([yshift=1.6mm]p.west);
\draw[predf] (p) -- (zh);
\draw[predf] (zh) -- (lp);
\draw[encf]  (zt1.east) -- (lp.west);
\draw[slate!150, dotted] (f1.south) -- (f2.north);
\node[tag, fill=white, inner sep=1.2pt, text=ink!75]
      at (2.10, 1.71) {shared $\theta$};

% ---------------- IDM block (training only) ----------------
\node[idm] (h) at (6.05,-0.36) {$h_\xi$\\[-1pt]{\scriptsize shared IDM head}};
\node[loss, align=center] (le) at (4.35,-1.62)
      {$\Lenc$\\[-1pt]{\scriptsize $h_\xi(z_t,z_{t+1})$ vs.\ $a_t$}};
\node[loss, align=center] (li) at (8.05,-1.62)
      {$\Lpid$\\[-1pt]{\scriptsize $h_\xi(\sgop(z_t),\hat z_{t+1})$ vs.\ $a_t$}};

\coordinate (ea) at (3.60,-0.10);
\coordinate (eb) at (4.55,-0.10);
\coordinate (sa) at (3.60,-0.66);
\coordinate (pa) at (8.85,-0.66);

\draw[encf, -] (zt.south)  -- (ea);
\draw[encf, -] (zt1.south) -- (eb);
\draw[encf]    (ea) -- ([yshift=2.6mm]h.west);
\draw[predf, -](ea) -- (sa);
\draw[predf]   (sa) -- ([yshift=-2.6mm]h.west);
\draw[predf, -](zh.east) -- ++(0.35,0) |- (pa);
\draw[predf]   (pa) -- ([yshift=-2.6mm]h.east);
\fill[teal!165] (eb) circle (1.1pt);
\fill[teal!165] (ea) circle (1.1pt);
\node[tag, anchor=north, text=indigo!160] at (4.45,-0.72) {$\sgop(z_t)$};
\node[tag, anchor=south, text=indigo!160] at (8.15,-0.60) {$\hat z_{t+1}$};

\draw[encf]  (h.south west) to[out=250,in=35]  (le.north east);
\draw[predf] (h.south east) to[out=290,in=145] (li.north west);

\draw[gradf] (le.west) -- ++(-0.35,0) |- (2.10,-1.62) -- (f2.south);
\node[tag, anchor=east, align=right, text=rose!175]
      at (1.95,0.30) {gradient\\shapes $f_\theta$};

\begin{scope}[on background layer]
  \node[draw=apricot!130, dashed, dash pattern=on 2pt off 1.5pt, fill=apricot!12, rounded corners=4pt, inner sep=3mm,
        fit=(h)(le)(li)(ea)(pa)] (idmbox) {};
\end{scope}
\node[tag, text=apricot!185, inner sep=1pt, align=center]
      at ($(idmbox.north)+(0,0.34)$) {training only,\\[-2pt]discarded at test time};

% ---------------- separator + test-time panel ----------------
\draw[slate!150, dashed] (9.95,3.30) -- (9.95,-2.55);
\node[anchor=north, font=\footnotesize\sffamily\bfseries] at (12.30,3.34) {Test time (planning)};

\node[enc, align=center]  (ta) at (11.15, 2.42) {$z_t=f_\theta(x_t)$};
\node[act, align=center]  (tb) at (13.45, 2.42) {CEM samples\\[-1pt]$g_\phi(a_{t:t+H_a})$};
\node[pred, align=center] (tc) at (12.30, 1.20) {$p_\psi$ rollout, $H_a$ steps};
\node[lat]  (td) at (12.30, 0.25) {$\hat z_{t+1:t+H_a}$};
\node[loss, align=center] (te) at (12.30,-0.80)
      {CEM cost\\[-1pt]$\tfrac{1}{H_a}\sum_{k}\lVert\hat z_{t+k}-z_\text{goal}\rVert^2$};
\node[box, fill=slate!35, draw=slate!150, align=center] (tf) at (12.30,-2.05)
      {$z_\text{goal}$: encoded goal frame,\\[-1pt]or language head (\S\ref{sec:nl})};

\draw[encf]  (ta.south) to[out=270,in=150] (tc.north west);
\draw[flow]  (tb.south) to[out=270,in=30]  (tc.north east);
\draw[predf] (tc) -- (td);
\draw[flow]  (td) -- (te);
\draw[flow]  (tf) -- (te);
\draw[flow, dashed, dash pattern=on 1.6pt off 1.3pt, draw=ink!60]
      (te.east) -- (14.72,-0.80) -- (14.72,2.42) -- (tb.east);
\node[tag, text=ink!70, rotate=90] at (14.92,0.80) {CEM iterates};

\end{tikzpicture}}
\caption{\textbf{\ours.} \emph{Left (training):} the \lewm backbone is unchanged --- encoder $f_\theta$, action encoder $g_\phi$, predictor $p_\psi$, latent prediction loss $\Lpred$. We add one shared inverse-dynamics head $h_\xi$ (orange), applied twice per batch with the same weights: to the encoder pair $(z_t,z_{t+1})$, whose gradient flows back into $f_\theta$ and forces consecutive latents to retain the action, and to the predictor pair $(\sgop(z_t),\hat z_{t+1})$, which forces the predicted latent to carry an action signature the same head can read. \emph{Right (test):} the head is discarded; CEM rolls out encoder and predictor under sampled actions and minimises the latent goal cost averaged over the rollout, with $z_\text{goal}$ from an encoded goal frame or, in \cref{sec:nl}, from the language head.}
\label{fig:architecture}
\end{figure}
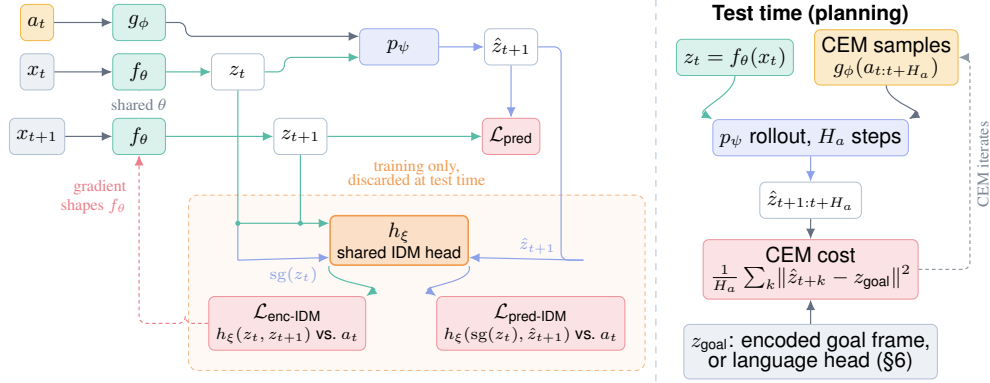

\paragraph{The loss.} We add one auxiliary objective: a shared inverse-dynamics head $h_\xi:\mathbb{R}^{2D}\to\mathbb{R}^{|a|}$, with $D$ the latent width, is a 3-layer MLP (width 256, $\sim167$K parameters) that predicts the frameskip-packed action from a pair of latents; it is applied twice per batch with the same weights.
\begin{align}
\Lenc &= \lVert h_\xi(z_t,z_{t+1})-a_t\rVert^2, \label{eq:enc}\\
\Lpid &= \lVert h_\xi(\sgop(z_t),\hat z_{t+1})-a_t\rVert^2. \label{eq:pid}
\end{align}
\emph{Encoder-IDM} (\cref{eq:enc}) back-propagates through $f_\theta$ and forces consecutive encoder latents to retain whatever content is needed to recover the action; it says nothing about the predictor. \emph{Predictor-IDM} (\cref{eq:pid}) requires the predicted latent to carry an action-decodable signature that the \emph{same} head can read, so the predictor cannot write action information into a private subspace the encoder does not occupy. $\sgop$ is a stop-gradient, so \cref{eq:pid} updates $p_\psi$ and $h_\xi$ but not $f_\theta$; the head is anchored by \cref{eq:enc} on real latent pairs (\cref{app:details}). The full objective is
\begin{equation}
\mathcal{L}_\text{I}=\Lpred+\lambda_\text{sigreg}\mathcal{L}_\text{sigreg}+\alpha_\text{inv}\Lenc+\alpha_\text{pid}\Lpid,
\end{equation}
with $(\lambda_\text{sigreg},\alpha_\text{inv},\alpha_\text{pid})=(0.09,1,1)$ held fixed for both benchmarks and every seed. The head is discarded at test time, so planning uses exactly the vanilla \lewm modules; the head adds under $1\%$ of the parameters. We call the result \ours, and the unrepaired baseline \emph{plain}.

\section{Experiments}
\label{sec:exp}

\subsection{Setup}
\label{sec:setup}
 Unless stated otherwise, encoder, predictor, optimiser and planner are identical across every cell and both benchmarks; only the I switch, the dataset and the seed vary. The encoder is a ViT-Tiny (5.5M parameters) trained \emph{from scratch}, the predictor a 6-layer transformer with context $H=3$, $\sim18$M trainable parameters in total; full hyper-parameters are in \cref{app:details}. We report the final training checkpoint (epoch 30). PushT uses the standard 10K-episode release \citep{chi2023diffusionpolicy}; actions are frameskip-packed (one predictor step $=$ 5 environment steps). The planner is CEM with action horizon $H_a\in\{5,10\}$ predictor steps ($H_a{=}5$ is the training horizon), replanning after each frameskip block on \bench and after the full plan on PushT, following PushT's published protocol and, on \bench, the cadence under which \ours plans best (\cref{app:cadence} swaps them). A \bench number is a \emph{trial mean}: 50 evaluation seeds $\times$ 4 tiers $\times$ 2 planner horizons, $N=400$ paired trials per model (200 scenes at two horizons), compared by exact McNemar tests between two trained models.

\subsection{Planning results}
\label{sec:results}

\begin{table}[t]
\centering\small
\setlength{\tabcolsep}{5pt}
\resizebox{\textwidth}{!}{%
\begin{tabular}{l cccc c cc}
\toprule
& \multicolumn{5}{c}{\bench (three training seeds, both horizons)} & \multicolumn{2}{c}{PushT} \\
\cmidrule(lr){2-6}\cmidrule(lr){7-8}
model & easy & medium & hard & very hard & trial mean & $H_a{=}5$ & $H_a{=}10$ \\
\midrule
\lewm (plain)  & 0.007 & 0.000 & 0.003 & 0.000 & $0.003\pm0.001$ & $0.98\pm0.01$ & $0.53\pm0.03$ \\
\ours (ours)   & \textbf{0.990} & \textbf{0.243} & \textbf{0.157} & 0.000 & $\mathbf{0.348\pm0.016}$ & $0.97\pm0.01$ & $\mathbf{0.83\pm0.01}$ \\
\midrule
\emph{random or zero-action policy} & 0.00 & 0.00 & 0.00 & 0.00 & 0.00 & --- & --- \\
\emph{goal-agnostic, one direction (tuned in hindsight)} & 0.12 & 0.20 & 0.00 & 0.00 & 0.08 & --- & --- \\
\emph{scripted controller, privileged state} & 1.00 & 1.00 & 1.00 & 0.28 & 0.82 & --- & --- \\
\bottomrule
\end{tabular}}

\caption{\textbf{Headline results.} Per-tier values pool three training seeds and both horizons (300 trials each); trial-mean and PushT columns are mean $\pm$ SEM over seeds (McNemar plain vs.\ \ours, canonical seed: $p<10^{-43}$; per-seed values in \cref{app:details}). Reference rows use the same scenes, budgets and thresholds: random or zero actions, the best single direction and duration chosen per tier in hindsight (it never reads the goal), and a scripted controller with simulator state (\cref{app:ceiling}).}
\label{tab:headline}
\end{table}

On \bench (\cref{tab:headline}) plain fails uniformly: 3 of 1200 trials across three training seeds. \ours reaches $0.348\pm0.016$; navigation saturates at $99\%$ and very hard floors at zero for both, so the pushing tiers carry the comparison. As evidence of goal-directed pushing we use the hard tier, where success rises from $0$ to $0.16$ and a policy that ignores the goal scores $0$; on medium \ours reaches $0.24$, but a single movement direction chosen in hindsight already reaches $0.20$ (\cref{sec:bench}), so medium says little about goal-directed pushing. A scripted controller with simulator state solves easy, medium and hard in full and very hard in $0.28$ of trials within the budget ($1.00$ with 60 steps; \cref{app:ceiling}), so the pushing ceiling lies in the learned model and its cost, and the very-hard zero partly in the budget. On PushT both models saturate at the trained horizon, but at twice that horizon plain drops to $0.53$ while \ours holds $0.83$ (no seed overlap).

\subsection{Ablations: which gradient does the work?}
\label{sec:disentangle}

\Cref{tab:disentangle} isolates which component carries the effect with ablation cells that share data and training budget.

\begin{table}[t]
\centering\small
\resizebox{\textwidth}{!}{%
\begin{tabular}{l cc c c l}
\toprule
cell & $\alpha_\text{inv}$ & $\alpha_\text{pid}$ & encoder & trial mean & vs.\ plain (McNemar) \\
\midrule
plain                       & 0 & 0 & ViT-Tiny, trained  & 0.000 & --- \\
predictor-IDM only          & 0 & 1 & ViT-Tiny, trained  & 0.010 & $+1.0$\pp ($p=0.13$) \\
encoder-IDM only            & 1 & 0 & ViT-Tiny, trained  & 0.290 & $+29.0$\pp ($p<10^{-34}$) \\
shuffled-action null        & 1 & 1 & ViT-Tiny, trained  & 0.003 & $+0.3$\pp ($p=1$) \\
frozen DINOv2-S, no I       & 0 & 0 & DINOv2-S, frozen   & 0.048 & $+4.8$\pp \\
frozen DINOv2-S + I         & 1 & 1 & DINOv2-S, frozen   & 0.163 & $+16.3$\pp$^\dagger$ \\
\ours (full)                & 1 & 1 & ViT-Tiny, trained  & \textbf{0.368} & $+36.8$\pp ($p<10^{-43}$) \\
\bottomrule
\end{tabular}}

\caption{\textbf{Disentangle.} Canonical training seed (the highest of three for \ours), on which McNemar pairs trials; three-seed means are $0.003$ (plain), $0.348$ (\ours), $0.298$ (encoder-IDM) and $0.002$ (shuffled null), per-seed values in \cref{app:details}. $^\dagger$Against frozen DINOv2-S without I: $+11.5$\pp, $p<10^{-9}$.}
\label{tab:disentangle}
\end{table}

\emph{Encoder-IDM is load-bearing.} Predictor-IDM alone yields $4/400$ successes, indistinguishable from plain (McNemar $p=0.13$). Encoder-IDM alone reaches $0.298\pm0.005$ over three training seeds, about $85\%$ of the full gain over plain (both three-seed); predictor-IDM adds the remaining $5$\pp on top of it (canonical seed $+7.8$\pp, McNemar $p<10^{-6}$, on medium and hard), so it is a complement that needs the encoder term, not a carrier on its own. \emph{The gradient, not the head, matters.} In the shuffled-action null the IDM targets are permuted across the minibatch (consistent latent pairs, wrong actions): both IDM losses stay at the action variance and the model stays at plain's level ($0.002\pm0.001$ over three seeds) at fixed architecture and compute. \emph{A stronger frozen encoder is not a substitute.} With a frozen DINOv2-S encoder \citep{oquab2024dinov2,zhou2024dinowm}, I lifts success from $0.048$ to $0.163$ but stays below half of \ours (easy $0.50$ vs.\ $1.00$): the gradient shapes the thin $384\to192$ projector, not the frozen backbone.

\subsection{Mechanism analysis}
\label{sec:mechanism}

\begin{figure}[t]
\centering
\includegraphics[width=0.78\textwidth]{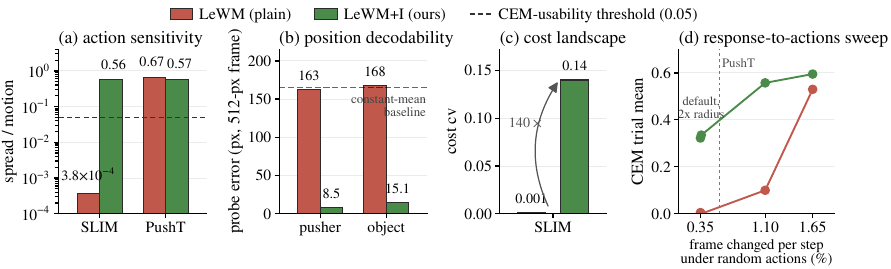}
\caption{\textbf{Mechanism probes} on \bench (and PushT for reference), \lewm vs.\ \ours. (a)~Plain's latent is almost action-insensitive, two orders of magnitude below the usability threshold and three below \ours; PushT-plain is action-sensitive without I. (b)~Pusher and object positions cannot be decoded from plain's latent (dashed line: constant mean-position predictor) and are recovered to within one patch by \ours; errors in the 512-px simulator frame. (c)~Plain's CEM cost surface is flat; \ours restores a usable descent direction. (d)~Planning success against the fraction of the frame that changes per step under random actions, over four environment variants (\cref{app:e1}): plain stays near zero until enough of the frame responds.}
\label{fig:mechanism}
\end{figure}
Four model-internal probes need no environment interaction. \textbf{Action sensitivity}: push $K=32$ random actions through the predictor from each held-out $z_t$ and report $\text{spread}/\text{motion}=\overline{\mathrm{Var}_k[\hat z^{(k)}_{t+1}]}\,/\,\overline{\lVert\hat z^{(k)}_{t+1}-z_t\rVert^2}$, the variance of the predicted latent across the $K$ actions over its mean squared displacement from $z_t$ (both per latent dimension, averaged over held-out contexts), a scale-free ratio that drift cannot fake. \textbf{Rollout error}: prediction error over a 10-step rollout divided by the true per-step latent motion; copying the latent forward scores on the order of the rollout length. \textbf{Position decodability}: pixel error of an MLP probe from $z_t$ to pusher and object positions. \textbf{Cost landscape}: coefficient of variation (cv) of the CEM cost over 1000 random action sequences from a start state, averaged over 200 start states (the 50 evaluation seeds $\times$ 4 tiers).

\Cref{fig:mechanism} (numbers in \cref{tab:probes}, appendix) summarises the probes. On \bench-plain the latent is action-insensitive (spread/motion $3.8\times10^{-4}$), the 10-step rollout error is $12.4\times$ the per-step true motion, on the order of copying the current latent forward (about the rollout length, 10), the pusher position is undecodable (163\,px, at the $\approx$165\,px of a constant mean-position predictor), and the cost surface is flat ($\text{cv}=0.001$). \ours restores all four: spread/motion $0.56$--$0.61$ across two seeds, rollout-error ratio $1.6$--$1.7$, pusher decodable to $8.5$\,px (within one patch), $\text{cv}=0.14$, and its latent cost ranks sampled actions like the simulator (Spearman $0.84$, \cref{app:spearman}). PushT-plain has none of the failure modes (spread/motion $0.67$; its predicted latent moves as far as the true one), consistent with plain planning well on PushT at the trained horizon.

\paragraph{Action sensitivity is necessary, not sufficient.} Across the twenty-eight configurations we trained (\cref{fig:threshold}, appendix), every one below spread/motion $\approx0.05$ fails, but above it success ranges from $0.01$ to $0.83$, and varying only the loss weight reproduces this inside one recipe (\cref{app:alpha}). The probe is a gate on plannability, not a ranking of planners.

\paragraph{The dose is how much of the frame responds to actions.} We vary it at fixed data volume and training budget (\cref{fig:mechanism}d; \cref{app:e1}) and measure it model-free, as the fraction of the frame that changes per predictor step under the environment's random actions. Doubling the object radius quadruples the footprint, but the objects still rarely move and the changing fraction stays at $0.36\%$ (default $0.35\%$): plain stays collapsed (0 of 400 trials, spread/motion $3\times10^{-4}$) and \ours plans as before ($0.335$ vs.\ $0.323$). Making the scene respond, with lighter, less damped objects and a larger, faster pusher, raises the fraction to $1.10\%$ and $1.65\%$: plain's encoder no longer collapses (spread/motion $0.88$ on both arms) and it starts to plan ($0.100$ and $0.530$), while \ours stays ahead ($0.558$ and $0.595$). PushT, where plain needs no help, lies at $0.57\%$. Within \bench, the vanilla encoder collapses only where too little of the frame responds; larger or faster-moving bodies can both raise that fraction, enlarging the objects alone did not, so what matters is the fraction rather than object size; the auxiliary loss supplies the action signal the frames lack. Distractor-free training, walls, ViT-Small and OGBench: \cref{app:crossdist,app:walls,app:vits,app:ogbench}.

\section{Language-conditioned goals}
\label{sec:nl}

\subsection{Language-goal head}
% TikZ: language-goal head pipeline + staged curriculum (Sec. 6)
\begin{figure}[t]
\centering
\resizebox{0.95\textwidth}{!}{%
\begin{tikzpicture}[
  font=\footnotesize\sffamily,
  >={Latex[length=1.7mm,width=1.5mm]},
  box/.style  ={draw, line width=0.4pt, rounded corners=2.5pt, align=center, inner sep=3pt,
                minimum height=6mm},
  img/.style  ={box, fill=slate!35, draw=slate!150, minimum width=8mm},
  enc/.style  ={box, fill=teal!35, draw=teal!150, minimum width=9mm},
  txt/.style  ={box, fill=lilac!45, draw=lilac!160},
  head/.style ={box, fill=apricot!60, draw=apricot!185, line width=0.7pt, text width=30mm,
                minimum height=13mm},
  lat/.style  ={box, fill=white, draw=slate!150, minimum width=9mm},
  pred/.style ={box, fill=indigo!30, draw=indigo!140, text width=33mm},
  loss/.style ={box, fill=rose!40, draw=rose!150, text width=31mm,
                minimum height=11mm},
  stage/.style={box, fill=slate!25, draw=slate!150, text width=29mm,
                minimum height=11mm, font=\scriptsize},
  flow/.style ={->, line width=0.5pt, rounded corners=3pt, draw=ink!85},
  tag/.style  ={inner sep=1.5pt, font=\scriptsize, text=ink!70}]

% ================= inference (top) =================
\node[font=\footnotesize\sffamily\bfseries, anchor=west] at (-1.05,3.35)
      {Inference: the language goal replaces the goal frame};

% visual branch
\node[img] (xt)   at (0.25,1.55) {$x_t$};
\node[enc] (f)    at (1.75,1.55) {$f_\theta$};
\node[lat] (zt)   at (3.80,1.55) {$z_t$};
\node[tag, below=0.5pt of f] {frozen};

% language branch
\node[txt, text width=30mm] (sent) at (-0.15,-0.35)
      {``push the red object\\[-2pt]to the green zone''};
\node[enc, minimum width=13mm] (mini) at (2.35,-0.35) {MiniLM};
\node[lat] (enl)  at (3.80,-0.35) {$e_\text{NL}$};
\node[tag, below=0.5pt of mini] {frozen};

% head + goal latent
\node[head] (h)   at (6.35,0.60) {$h_\Pi$\\[-2pt]language-goal head\\[-1pt]
                                  {\scriptsize cross-attention over\\[-3pt]patch tokens, trained}};
\node[lat]  (zg)  at (8.70,0.60) {$z_\text{goal}$};

% planner
\node[pred] (roll) at (7.45,2.70) {$p_\psi$ rollout under\\[-1pt]CEM action samples};
\node[loss] (cost) at (10.85,1.55) {CEM cost\\[-1pt]$\tfrac{1}{H_a}\sum_{k}\lVert\hat z_{t+k}-z_\text{goal}\rVert^2$};
\node[tag, below=0.5pt of roll] {frozen};

% --- arrows ---
\draw[flow] (xt)   -- (f);
\draw[flow] (f)    -- (zt);
\draw[flow] (sent) -- (mini);
\draw[flow] (mini) -- (enl);

\draw[flow] (zt.south)  -- (3.80,0.95) -- ([yshift=3.5mm]h.west);
\draw[flow] (enl.north) -- (3.80,0.25) -- ([yshift=-3.5mm]h.west);
\draw[flow] (h) -- (zg);

\draw[flow] (zt.north)   -- (3.80,2.70) -- (roll.west);
\draw[flow] (roll.east)  -- (10.85,2.70) -- (cost.north);
\draw[flow] (zg.east)    -- (10.85,0.60) -- (cost.south);

% ================= training (bottom) =================
\draw[slate!150, dashed] (-1.05,-1.55) -- (12.5,-1.55);
\node[font=\footnotesize\sffamily\bfseries, anchor=west] at (-1.05,-1.95)
      {Training: predict the next stage from the current frame, world model frozen};
\node[stage] (s1) at (0.60,-3.00) {\textbf{random pose}\\three per scene,\\before contact};
\node[stage] (s2) at (4.05,-3.00) {\textbf{approach}\\``Position the pusher behind the red object, aligned with the green zone''};
\node[stage] (s3) at (7.50,-3.00) {\textbf{midway}\\``Push the red object halfway toward the green zone''};
\node[stage] (s4) at (10.95,-3.00) {\textbf{goal}\\``The red object is at the green zone''};
\draw[flow, draw=ink!60] (s1) -- (s2);
\draw[flow, draw=ink!60] (s2) -- (s3);
\draw[flow, draw=ink!60] (s3) -- (s4);
\node[anchor=north west, font=\footnotesize, text=ink!90, align=left, text width=13.3cm, inner sep=0pt] at (-1.05,-3.72)
      {each arrow is one training pair: (latent of the left frame, sentence of the right stage) $\to$ latent of the right stage; the goal also maps to itself; 
       40\,000 scenes $\times$ 6 frames.};
\end{tikzpicture}}
\caption{\textbf{Language-goal head.} \emph{Top (inference):} a small trained head maps the frozen encoder's $z_t$ and a frozen sentence embedding of the goal description to a goal latent $z_\text{goal}$, which replaces the encoded goal frame in the CEM cost; encoder, predictor and planner are unchanged. \emph{Bottom (training):} from each frame the head predicts the encoded latent of the next stage of a push, given that stage's sentence; at test time it is always asked for the final goal.}
\label{fig:langhead}
\end{figure}
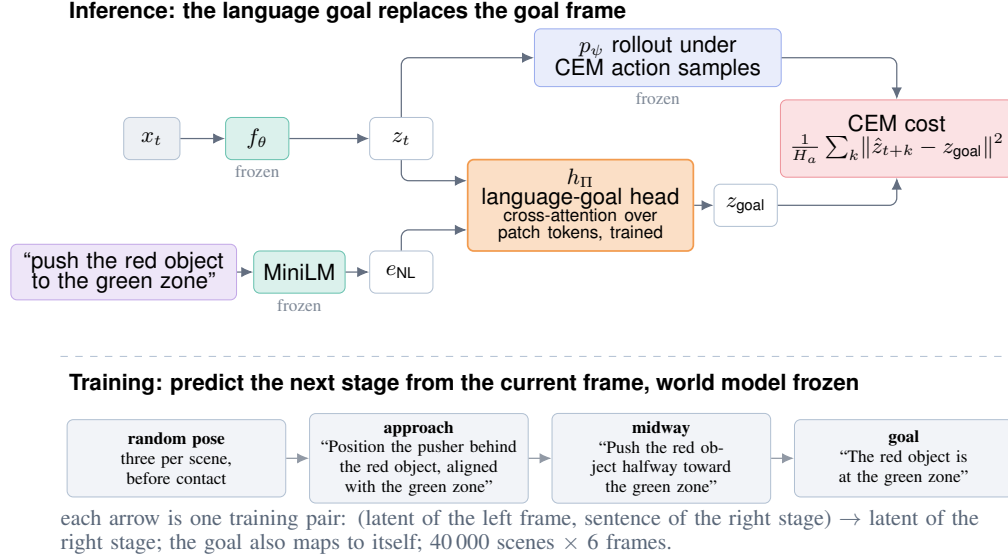

\paragraph{Head.} A small head $h_\Pi(z_t,e_\text{NL})\mapsto z_\text{goal}$ (\cref{fig:langhead}) produces the goal latent for the planner's cost; $e_\text{NL}$ is the embedding of the goal sentence from a frozen \texttt{all-MiniLM-L6-v2} encoder. Two cross-attention layers let a query from $z_t$ and one from $e_\text{NL}$ attend over the encoder's 256 patch tokens (1.1M parameters, \cref{app:ca}). The head is trained on $40\,000$ simulator scenes, disjoint from the evaluation seeds, each rendered as six frames along a push with one sentence per stage (\cref{fig:langhead}); from each frame it predicts the encoded latent of the \emph{next} stage. In the main evaluation it receives the final goal sentence (``The red object is at the green zone''), a pairing with the start frame it never saw in training; \cref{sec:staged} instead gives it the stage sentences it was trained on. It is queried twice per 20-step episode with the same sentence, at the start and again after 10 planner steps from the then-current latent, so the goal latent is refreshed from the new state; the planner uses the latest output in between. Evaluation withholds the goal frame; the visual-goal planner of \cref{tab:headline} on the same world model is the \emph{oracle}.

\begin{table}[t]
\centering\small
\setlength{\tabcolsep}{4.5pt}
\resizebox{\textwidth}{!}{%
\begin{tabular}{l cccc c l}
\toprule
goal channel & easy & medium & hard & very hard & trial mean & $\Delta$ vs.\ oracle (McNemar) \\
\midrule
\emph{goal-agnostic, one direction (hindsight)} & 0.12 & 0.20 & 0.00 & 0.00 & 0.08 & --- \\
visual-goal oracle                     & 1.00 & 0.30 & 0.17 & 0.00 & 0.368 & --- \\
goal-frame-conditioned head (MLP)      & 0.99 & 0.29 & 0.17 & 0.00 & 0.363 & $-0.5$\pp ($p=0.80$) \\
language head over the global latent (MLP) & 0.53 & 0.16 & 0.09 & 0.00 & 0.195 & $-17.3$\pp ($p<10^{-11}$) \\
\textbf{language head, patch attention, 3 seeds} & \textbf{0.84} & 0.06 & 0.04 & 0.00 & $\mathbf{0.235\pm0.026}$ & $-13.3$\pp ($p<10^{-9}$ each seed) \\
\midrule
\quad zero sentence embedding          & 0.00 & 0.00 & 0.01 & 0.00 & 0.003 & \multicolumn{1}{l}{vs.\ head: $p<10^{-27}$} \\
\quad wrong colour word                & 0.10 & 0.00 & 0.00 & 0.00 & 0.025 & \multicolumn{1}{l}{vs.\ head: $p<10^{-24}$} \\
\quad target fixed at $t{=}0$          & 0.74 & 0.00 & 0.04 & 0.00 & 0.195 & \multicolumn{1}{l}{vs.\ head: $p<10^{-3}$} \\
\quad paraphrases, train pool          & 0.84 & 0.04 & 0.02 & 0.00 & 0.225 & \multicolumn{1}{l}{vs.\ head: $p=0.15$} \\
\quad paraphrases, unseen nouns        & 0.78 & 0.01 & 0.03 & 0.00 & 0.205 & \multicolumn{1}{l}{vs.\ head: $p=0.005$} \\
\bottomrule
\end{tabular}}

\caption{\textbf{Language goals} on the \ours world model of the canonical seed, whose visual-goal planner is the oracle (per-tier values average both horizons; $\pm$: SD over head seeds). Indented rows ablate head seed 42 ($0.245$), against which their p-values are computed; per-seed values in \cref{app:details}.}
\label{tab:nlhead}
\end{table}

\subsection{Results}
\label{sec:nlresults}
\paragraph{The head grounds the named zone.} Attending over the encoder's patch tokens, the head reaches $0.84$ on navigation against the oracle's $1.00$ and $0.235\pm0.026$ over all tiers (\cref{tab:nlhead}). It depends on the sentence (zeroing its embedding leaves $1/400$), and naming a colour that appears nowhere in the scene drops it to $0.025$. To test whether it reads the colour or has learned where targets usually lie relative to the pusher, we swap the named zone with the other zone on the navigation tier, so that a zone of the wrong colour sits at the typical target position. The head never ends at this decoy, ends closer to the named zone in all 300 episodes (50 scenes, 3 seeds, 2 horizons), and succeeds as often as without the swap ($252$ vs.\ $253$): it goes where the sentence says, not where targets usually are. An MLP head conditioned on the goal frame instead of the sentence matches the oracle ($0.363$ vs.\ $0.368$, $p=0.80$), so the training pipeline and planner are not the bottleneck; the gap lies in mapping a sentence, which specifies a set of goal states, to one goal latent. An MLP language head over $z_t$ trades navigation ($0.53$) for pushing (medium $0.16$, hard $0.09$) and is less reliable under the swap.

\paragraph{Language degrades gracefully under paraphrase.} On seed 42 the head keeps $92\%$ of its success on 107 paraphrases ($p=0.15$) and $84\%$ on 31 held-out templates with unseen object and zone nouns ($p=0.005$), mostly on navigation. One query at $t{=}0$ already reaches $0.195$; a mid-episode refresh adds $5$\pp ($p<10^{-3}$).

\paragraph{Pushing from one goal sentence is the hard part.} On the pushing tiers the goal sentence reaches about a fifth of the oracle (medium $0.06$, hard $0.04$ vs.\ $0.30$ and $0.17$) and stays below the goal-agnostic policy on medium: from the start frame, a single sentence has to specify where a push ends that the planner cannot reach in one step. \Cref{sec:staged} shows that stage sentences close this gap to the goal-frame oracle.

\subsection{Staged instructions}
\label{sec:staged}
The head is trained on stage transitions, from a frame and the sentence of the next stage to that stage's latent, yet the evaluation above hands it the final goal sentence from the start frame. We now give it the push the way it was trained, as a sequence of stage sentences, the form of step-by-step instruction that a user or a language-model planner \citep{huang2022zeroshot,ahn2022saycan} can supply: ``Position the pusher behind the red object, aligned with the green zone'', then ``Push the red object halfway toward the green zone'', then ``The red object is at the green zone'' (navigation: halfway toward the zone, then the goal). The stage sentences name the same object and zone as the goal sentence; what they add is how the push proceeds. The head is queried with the next sentence every 7 planner steps (steps 0, 7 and 14 of 20), each time from the current frame. The oracle counterpart hands the planner the encoded frames of the same stages on the same schedule: pusher behind the object, object halfway, object in the zone (\cref{app:staged}). All rows of \cref{tab:staged} come from one evaluation campaign on one machine and pair the same 400 trials; the goal-frame oracle and goal-sentence rows therefore differ slightly from \cref{tab:nlhead} ($0.353$ vs.\ $0.368$, $0.229$ vs.\ $0.235$), which was evaluated on different hardware.

\begin{table}[t]
\centering\small
\setlength{\tabcolsep}{4.5pt}
\resizebox{\textwidth}{!}{%
\begin{tabular}{l cccc c l}
\toprule
subgoals & easy & medium & hard & very hard & trial mean & paired test (McNemar) \\
\midrule
goal frame (visual-goal oracle)          & 0.98 & 0.27 & 0.16 & 0.00 & 0.353 & --- \\
stage frames (oracle)                    & 0.99 & 0.62 & 0.55 & 0.01 & 0.543 & vs.\ goal frame: $p<10^{-18}$ \\
\midrule
goal sentence, 3 head seeds              & 0.84 & 0.04 & 0.04 & 0.00 & $0.229\pm0.021$ & --- \\
\textbf{stage sentences}, every 7 steps  & 0.82 & \textbf{0.25} & \textbf{0.25} & 0.01 & $\mathbf{0.334\pm0.059}$ & vs.\ goal sentence: $p<0.002$ each seed \\
\textbf{stage sentences}, latent switch  & 0.82 & \textbf{0.29} & \textbf{0.23} & 0.00 & $\mathbf{0.335\pm0.037}$ & vs.\ goal sentence: $p<10^{-4}$ each seed \\
stage sentences, paraphrases, train pool & 0.78 & 0.24 & 0.18 & 0.01 & $0.305\pm0.055$ & vs.\ goal sentence: $p<10^{-4}$, $0.08$, $<10^{-4}$ \\
stage sentences, paraphrases, unseen nouns & 0.74 & 0.09 & 0.06 & 0.00 & $0.223\pm0.012$ & vs.\ goal sentence: $p>0.2$ each seed \\
\midrule
\quad stage sentences, every 5 steps     & 0.82 & 0.25 & 0.23 & 0.02 & 0.330 & vs.\ every 7 steps: $p=0.69$ \\
\quad stage sentences, simulator-judged switch & 0.74 & 0.04 & 0.08 & 0.00 & 0.215 & vs.\ every 7 steps: $p<10^{-7}$ \\
\quad goal sentence, 30 planner steps    & 0.88 & 0.04 & 0.05 & 0.00 & 0.243 & --- \\
\quad stage sentences, 30 planner steps (every 10) & 0.85 & 0.28 & 0.25 & 0.00 & 0.345 & vs.\ goal sentence: $p<10^{-6}$ \\
\bottomrule
\end{tabular}}
\caption{\textbf{Staged instructions} on the \ours world model of \cref{tab:nlhead}; the language rows use the cross-attention head (per-tier values average both horizons and the three head seeds; $\pm$: SD over head seeds). Indented rows use head seed 42 (stage sentences every 7 steps: $0.340$). All rows come from one evaluation campaign on one machine; per-seed values in \cref{app:staged}.}
\label{tab:staged}
\end{table}

\begin{figure}[t]
\centering
\includegraphics[width=0.78\textwidth]{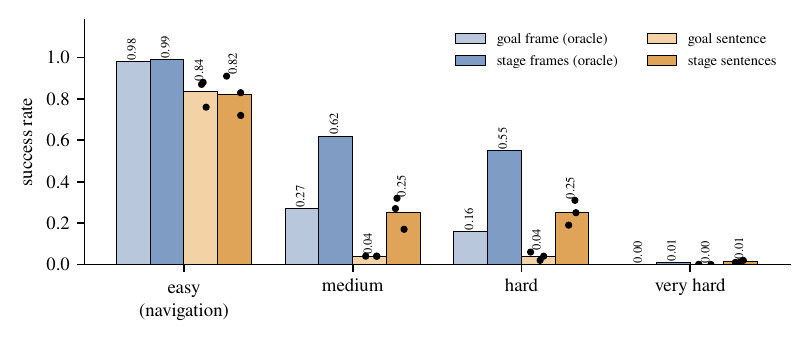}
\caption{\textbf{Goal vs.\ stage subgoals, as frames and as sentences.} Stage frames lift the planner on the pushing tiers far above the goal frame; stage sentences lift the language head from $0.04$ to the level of the goal-frame oracle there. Bars: means; dots: the three head seeds.}
\label{fig:staged}
\end{figure}

\paragraph{Stage sentences make pushing reachable.} Given the stage sentences, the head raises success on medium and hard from $0.04$ to $0.25$ and the trial mean from $0.229\pm0.021$ to $0.334\pm0.059$ (\cref{tab:staged,fig:staged}). The gain holds for every head seed (McNemar $p=2\times10^{-6}$, $1.1\times10^{-3}$ and $3\times10^{-13}$) and lies entirely on the pushing tiers; navigation is unchanged. On medium the head now reaches the goal-frame oracle on average ($0.25$ vs.\ $0.27$), and on hard it lies above it ($0.25$ vs.\ $0.16$), significantly for head seed 11 ($0.31$, $p=0.006$); on the pushing tiers no head seed is significantly below the oracle. Longer episodes do not substitute for the stages: with 30 planner steps (stages every 10 steps) the goal sentence stays at $0.243$, the stage sentences reach $0.345$.

\paragraph{Decomposition helps the planner itself.} The stage-frame oracle shows where the gain comes from. Handing the planner the three stage frames instead of the goal frame lifts medium from $0.27$ to $0.62$ and hard from $0.16$ to $0.55$ ($0.543$ vs.\ $0.353$, $p=2\times10^{-19}$): a nearby subgoal is easier for the planner to reach than the end of a push, consistent with the compounding rollout error that limits the very-hard tier (\cref{app:cadence}). The language head recovers $40$--$45\%$ of this oracle on pushing, so the remaining gap lies in its stage latents rather than in the planner. Queried from the rendered frame of the previous stage, the head's output lies nearest to the intended stage frame in $94$--$100\%$ of the pushing queries for head seeds 42 and 11 and in $76$--$100\%$ for seed 7, the weakest seed in closed loop (\cref{app:staged}); in closed loop it is queried from whatever frame the planner reached.

\paragraph{The switch needs neither a schedule nor the simulator.} Seven steps per stage is not a tuned value: switching every five steps gives the same result ($0.330$ vs.\ $0.340$, $p=0.69$). The schedule can also be dropped. Checking at every step and advancing once the latent distance to the current stage target has fallen to a tenth of its value at stage entry, or has not reached a new minimum for three steps, gives $0.335\pm0.037$, not significantly different from the fixed schedule for any head seed ($p=0.29$, $0.31$ and $0.06$); the two thresholds were chosen on separate tuning scenes (20 per tier), and the rule reads nothing but the world model's latent. A simulator judge that advances only once a stage is complete is worse ($0.215$, $p<10^{-7}$): most pushes never register the midway stage as complete and so never receive the goal sentence. Moving on when progress stalls matters more than knowing when a stage is done.

\paragraph{Grounding holds; new vocabulary does not yet.} With stage sentences the head still goes to the named zone in the decoy test: it never ends at the decoy and ends closer to the named zone in all 300 episodes, succeeding on $0.85$ of them (goal sentence in the same campaign: $0.88$). With paraphrases from the train pool, which reword the training templates in the same vocabulary, most of the pushing gain remains (medium $0.24$, hard $0.18$; trial mean $0.305\pm0.055$, above the goal sentence with $p<10^{-4}$ for head seeds 42 and 11 and $p=0.08$ for seed 7), although seed 11 keeps less of it than with the canonical sentences. With held-out paraphrases, whose object and zone nouns never occur in training and whose verbs overlap only partly, the pushing gain largely disappears (medium $0.09$, hard $0.06$; over all tiers no seed differs significantly from the goal sentence), while navigation keeps most of its success. Staged pushing thus largely survives rewording within the training vocabulary, but not new vocabulary.

\section{Conclusion}
\label{sec:discussion}
When little of the frame responds to actions, a JEPA world model collapses before language enters: an encoder that discards pusher and objects still meets the prediction objective. One inverse-dynamics auxiliary loss on encoder and predicted latents restores a plannable latent at no test-time cost, our controls attribute the repair to the gradient into the encoder, and the action-sensitivity probe offers a cheap check before planning. On the repaired latent a small head plans from sentences, approaches the goal-frame oracle on navigation, goes to the zone the sentence names even when a decoy occupies the usual target position, and, given a push as a sequence of stage sentences, completes pushes as often as the goal-frame oracle, with the switch between stages read from the latent.

\bench makes the next steps measurable: closing the gap between stage sentences and stage frames, making staged pushing robust to unseen wording, letting a language model decompose a goal into stage sentences, and carrying the repair to larger encoders, whose probes recover but whose planning did not yet transfer.

\subsection*{AI use statement}
In this work, we used generative AI tools (LLM-based coding assistants) to assist in implementing methods, probes and evaluation code, to clean and reformat datasets (memory-mapped conversion of the rendered episodes) and to provide feedback on research methodology and experiments (including suggesting control experiments and the response-to-actions sweep). We have not used generative AI tools to generate datasets: all training and evaluation data are produced by the seeded physics simulator of \cref{sec:bench}. Developing theoretical models, formulating or proving mathematical claims, translation, and qualitative or thematic data analysis are not applicable to this work. Additionally, we used generative AI tools for drafting and editing parts of the paper, creating and modifying figures and plotting scripts, searching and summarising related literature, formatting references, and proposing titles. We have reviewed all AI-assisted work: code was checked by the authors and validated throughout, every reported number was traced to an evaluation log, and AI-suggested related work was verified against the original papers. We take responsibility for the final content of this work, including text, claims or artifacts produced with the aid of generative AI.

\subsection*{Ethics statement}
This work uses only synthetic 2D simulation data and public benchmark releases; it involves no human subjects, personal data or deployment. We see no direct negative societal impact beyond those generic to robot learning.

\subsection*{Reproducibility statement}
\Cref{sec:bench} and \cref{app:details} specify the benchmark (simulator constants, tier generator, success thresholds, the seeded data-collection policy and the train/validation split), the model and optimiser, and the planner protocol (path cost, 20 planner steps, 50 fixed evaluation seeds per tier shared by every cell); \cref{app:traindist} characterises the training data, \cref{app:ceiling} the scripted reference controller, \cref{app:e1} the physics overrides of the response sweep, \cref{app:ca,app:layout} the language head, its curriculum, its paraphrase pools and the grounding controls, and \cref{app:staged} the staged protocol. Per-seed values, the statistical tests and their pairing are in \cref{app:details}. The code (environment, seeded dataset generator, training and evaluation code, probes, language templates and paraphrase pools) will be released.

\bibliography{main}
\bibliographystyle{iclr2027_conference}

\appendix
\section{Additional experimental details}
\label{app:details}

\paragraph{Environment constants.} \bench is simulated at $512\times512$ internal resolution and rendered at $224\times224$. Pusher: radius 14\,px, mass 5, PD gains $K_p{=}100$, $K_v{=}20$, action scale 40 (calibrated to PushT's per-step pixel displacement), velocity cap 100\,px/s; the cap binds for action norms above about $0.45$, where the pusher moves 6--6.6\,px per environment step and the action mainly sets the direction of motion. Objects (1--3 per scene, 0--2 on the navigation tier; two zones per scene): radius 15\,px, mass 8, global damping 0.5, per-substep object damping 0.85. One environment step is ten physics substeps at 10\,Hz; one control (planner) step is five environment steps (frameskip). Tier configurations place pusher, target object and target zone by the tier's template, translate the template uniformly within the arena (all three at least 80\,px from the border), draw each distractor object and zone as the corresponding point of an independent, translated and rotated template of the same tier (rejecting overlaps), and then map the whole scene through one of the eight symmetries of the square arena, drawn per (seed, tier) from a separate random stream. On medium and hard the target object is then the most central object about as often as chance ($0.44$ and $0.43$ vs.\ $0.42$ and $0.41$ over 500 seeds); on very hard, whose template spans the arena diagonal and leaves little room for translation, it is the most central in $0.60$ of scenes (chance $0.42$), a residual cue on a tier that no planner solves. Walls, when enabled, are chains on a $40\times40$ grid with extension probability 0.9 and decay 0.1, traversability guaranteed. Zero action produces exactly zero pusher motion, which the action-sensitivity probe relies on. The environment's own termination threshold is 25\,px; evaluation success uses 30\,px (pusher) and 40\,px (object) to account for residual end-state distance after a closed-loop rollout, identically for every model.

\paragraph{Planner.} All PushT training and evaluation uses the stable-worldmodel platform \citep{maes2026stableworldmodel}. CEM with 30 iterations $\times$ 300 samples and top-30 elites on PushT, 20 iterations $\times$ 200 samples and top-20 elites on \bench, variance scaling 1.0, no momentum; the cost is the squared distance to the goal latent averaged over the $H_a$ predicted latents of the rollout (path cost); 20 planner steps per episode, each executed as five environment steps.

\paragraph{Success semantics.} Success is scored on the final step of a fixed-length episode. Scoring ``ever within threshold'' (checked after each executed block) instead raises \ours by $+1.3$\pp on the easy/medium/hard cells ($147\to151$ of 300 paired trials, discordant pairs $0/4$, McNemar $p=0.125$; three on medium, one on hard); per-episode minimum distances are stored, so either reading is reproducible.

\paragraph{Data.} 10\,000 episodes; training clips are split 90/10 at random into training and validation, identically for every variant (clips of one episode can fall on both sides, so the validation loss is a training diagnostic rather than a held-out measure); frames are resized to 224\,px and normalised as in LeWM, without augmentation, and no data-side choice is modified for \ours. Training episodes contain 3--6 objects and 1--3 zones (evaluation scenes: 1--3 objects, two zones). Inter-object pushes, in which the pusher moves one object that displaces another, occur in about $8\%$ of frames.

\paragraph{Model.} Encoder: ViT-Tiny, 192 hidden, 12 layers, 3 heads, patch 14 ($16\times16$ tokens at 224\,px), random initialisation. Predictor: 6 layers, hidden 192, 16 heads, MLP 2048, dropout 0.1, context $H{=}3$, supervised at all context positions. Action encoder: linear $\mathbb{R}^2\to192$. IDM head: MLP $384\to256\to256\to10$ (frameskip 5 $\times$ 2-dim actions). AdamW lr $5\times10^{-5}$, weight decay $10^{-3}$, gradient clipping 1.0, batch 128, bf16. Training 30 epochs; the final checkpoint (epoch 30) is evaluated, without peak-epoch selection; the one exception is a plain seed that diverged (below). A \bench evaluation of 20 episodes takes 3--4 minutes on one A100; a 50-episode PushT evaluation about 8 minutes.

\paragraph{Inverse-dynamics head.} The encoder-side term is applied to every consecutive latent pair of a training window (three pairs per window at context 3), the predictor-side term to the last context position only (one pair per window); both use unit weight, and the head is updated by both (the stop-gradient in \cref{eq:pid} is on $z_t$ only). In the shuffled-action null the latent pairs are kept and the action targets are permuted across the minibatch (for the encoder term, across minibatch and time steps), so the head receives inconsistent targets; the null therefore bounds what head capacity and compute contribute, and a permuted target can at most push the encoder away from action information. The training seed fixes the train/validation split; parameter initialisation and dropout are not tied to it, so repeated runs with the same seed value differ in initialisation.

\paragraph{Language head training.} AdamW, MSE $+0.1\times$ cosine loss, 90 epochs for the cross-attention head and 30 for the MLP head; the world model is frozen and the training scenes are encoded once.

\paragraph{Per-seed headline values.} \bench trial mean: \ours $0.368$ / $0.360$ / $0.315$ (training seeds 3072, 9999, 42), plain $0$ / $2$ / $1$ of 400 (seeds 3072, 1234, 5678). Plain seed 5678 diverged to non-finite weights in its 28th training epoch; we report its last finite checkpoint (epoch 27). McNemar tests pair the 400 evaluation trials (identical scenes and seeds) of two models; training seeds need not coincide; PushT $H_a{=}10$ (seeds 3072, 1234, 5678 for both models): plain $0.58$ / $0.52$ / $0.48$, \ours $0.82$ / $0.84$ / $0.82$. Encoder-IDM only: $0.290$ / $0.308$ / $0.298$ (seeds 3072, 9999, 42); shuffled-action null: $0.003$ / $0.000$ / $0.003$ (seeds 3072, 9999, 42). Language head (\cref{tab:nlhead}): trial means $0.245$ / $0.205$ / $0.255$ (head seeds 42, 7, 11; easy $0.88$ / $0.78$ / $0.87$), McNemar against the visual-goal oracle $p=1.5\times10^{-12}$, $5.1\times10^{-17}$ and $2.1\times10^{-10}$.

\begin{figure}[h]
\centering
\includegraphics[width=0.5\textwidth]{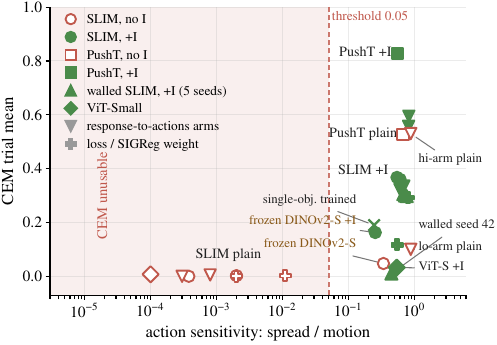}
\caption{Action sensitivity vs.\ CEM success for 28 trained configurations (circles \bench, squares PushT, triangles walled \bench, diamonds ViT-Small, downward triangles the response-to-actions arms, plus signs the loss-weight and SIGReg cells; filled with I, hollow without; the cross is \ours trained on the single-object variant and evaluated cross-distribution). Below $0.05$ every planner fails; above it success ranges from $0.01$ to $0.83$: the frozen-DINOv2 cells, the non-converged walled seed and ViT-Small clear the threshold yet plan poorly.}
\label{fig:threshold}
\end{figure}

\paragraph{Probe values.} \Cref{tab:probes} lists the probe values behind \cref{fig:mechanism}.

\begin{table}[h]
\centering\small
\begin{tabular}{l cc cc}
\toprule
probe & \bench plain & \bench \ours & PushT plain & PushT \ours \\
\midrule
action sensitivity, spread/motion     & $3.8\times10^{-4}$ & $0.56$ / $0.61$ & $0.67$ & $0.57$ \\
rollout error / true motion ($h{=}10$) & $12.4$ & $1.69$ / $1.62$ & $1.02^\dagger$ & $0.95^\dagger$ \\
pusher position MLP error (px)        & $163\pm2$ & $8.5\pm0.1$ & --- & --- \\
object position MLP error (px)        & $168\pm0.4$ & $15.1\pm0.2$ & --- & --- \\
cost-landscape cv                     & $0.001$ & $0.14$ & --- & --- \\
\bottomrule
\end{tabular}
\caption{Mechanism probes. $^\dagger$For PushT only the predicted-over-true latent displacement was measured, not the error ratio, so these two entries are not comparable with the \bench column. Two values for \bench \ours are two training seeds (within $10\%$ on spread/motion, $5\%$ on the rollout-error ratio). Position errors (mean $\pm$ SEM over three probe seeds, 80/10/10 split) are in the 512-px simulator frame, where a constant mean-position predictor scores $\approx165$\,px.}
\label{tab:probes}
\end{table}

\paragraph{Other auxiliary objectives.} We also trained two other auxiliaries on \bench: a residual predictor (R) and a patch decoder (P), alone and combined. Every cell without I kept plain-level probes (spread/motion $5.8\times10^{-3}$, $1\times10^{-4}$ and $6\times10^{-4}$ for R, P and R+P; pusher decodability $166$\,px), whereas every cell with I restored them (I $0.56$, I+P $0.54$, R+I+P $0.70$ spread/motion; pusher decodability $8.5$, $7.4$ and $5.3$\,px). P is a small refinement on top of I and R is redundant, so the paper uses I alone.

\subsection{Planner cadence control}
\label{app:cadence}
The benchmarks inherit different receding-horizon cadences from their upstream pipelines. Running \bench with PushT's full-plan cadence lowers \ours from $0.490$ to $0.390$ trial mean on the six easy/medium/hard cells ($147/300\to117/300$, McNemar $p=1.4\times10^{-6}$; losses concentrated on medium and hard); running PushT with \bench's replan-every-block cadence crashes both models ($H_a{=}10$: \ours $0.83\to0.20$, plain $0.53\to0.18$). Each benchmark is most usable under its native cadence, the mismatch is symmetric, and the \ours-vs-plain gap keeps its sign under both.

\paragraph{Horizon and cost mode.} On PushT the I advantage vanishes at $H_a=15$ (both $0.04$), three times the trained horizon. Path cost (the mean over the $H_a$ predicted latents) beats terminal cost: the r=30 checkpoint of \cref{app:e1} scores $0.335$ under path cost and $0.083$ under terminal cost over 400 paired trials (McNemar $p<10^{-21}$). Compounding predictor error limits the very-hard tier; within \ours the cost-landscape cv is $0.16$--$0.17$ on easy, medium and hard but $0.07$ on very hard.

\section{Robustness to obstacle walls}
\label{app:walls}
\begin{figure}[h]
\centering
\includegraphics[width=0.55\textwidth]{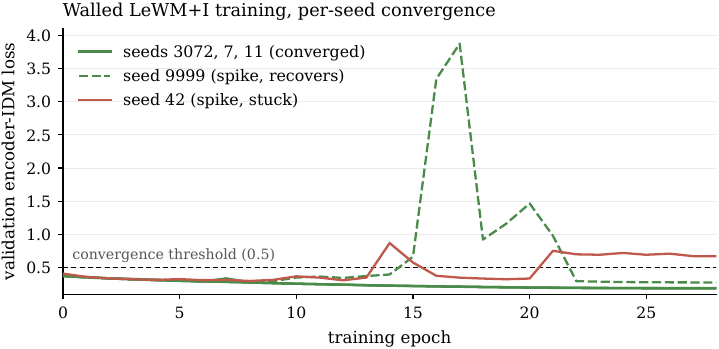}
\caption{Validation encoder-IDM loss per epoch for the five walled \ours seeds; seed 9999's spike (peak $3.9$ at epoch 17) exceeds the axis range.}
\label{fig:walled}
\end{figure}

We regenerated the dataset with procedural walls (10K episodes; the wall generator turns chains toward increasing coordinates in $82\%$ of turns) and retrained plain (one seed) and \ours (five seeds) for 30 epochs. Walled plain (seed 3072): $2/400$. Walled \ours: $0.310$, $0.298$, $0.300$, $0.305$ (seeds 3072, 9999, 7, 11; McNemar vs.\ plain $p<10^{-30}$ each) and $0.010$ (seed 42). Conditional on convergence the four seeds cluster at $0.303\pm0.006$, close to the open arena's pooled $0.348$; unconditionally $0.245\pm0.131$ with a 1-in-5 failure (this and the other $\pm$ values in this appendix are SDs across seeds) (Jeffreys 95\% interval on the failure rate $[1.3\%,62\%]$). Convergence is defined by a training-side criterion fixed before evaluation: validation encoder-IDM loss below 0.5 at the last epoch (converged seeds 0.19--0.28, seed 42: 0.67). Two of five walled runs show a validation encoder-IDM spike, at epochs 14 and 21 (seed 42) and at epoch 17 (seed 9999, peak $3.9$); one recovers, one does not (\cref{fig:walled}). Per tier over the four converged seeds (mean $\pm$ SD): easy $0.90\pm0.02$, medium $0.27\pm0.03$ (open arena $0.24$), hard $0.04\pm0.02$, very hard $0$; hard drops because wall chains obstruct the steered pushes. Each McNemar test contrasts one \ours training instance with one plain instance on the same 400 trials, so it tests that pair of models, not the recipe distribution across training seeds.

\section{Encoder scale: ViT-Small}
\label{app:vits}
Retraining both models at ViT-Small (22M encoder parameters) with otherwise identical settings and the same 30-epoch budget preserves the probe asymmetry: plain spread/motion $2.0\times10^{-4}$ and rollout-error ratio $28.4$; \ours $0.525$ and $2.08$. The planner result, however, degrades: \ours reaches only $0.033$ trial mean (plain $0.008$), and object-position decodability worsens from 15\,px to 72\,px, while the cost-landscape cv stays in the same range ($0.11$ vs.\ $0.14$ at ViT-Tiny, averaged over tiers), the sharpest instance of the probe's insufficiency. The comparison is at matched epochs, not matched convergence; we claim only that the mechanism is not a ViT-Tiny artefact at the representation level, and that the absolute success ceiling at our budget is ViT-Tiny-specific.

\section{OGBench cube: an expected null}
\label{app:ogbench}
On OGBench visual cube-single \citep{park2024ogbench} ($64\times64$ frames bilinearly upsampled to $224$, 1M frames, 30 epochs, $N{=}50$), plain and \ours are indistinguishable ($4/50$ vs.\ $3/50$, disjoint episodes, McNemar $p=1.0$), consistent with the single-object null on PushT at the trained horizon. On cube-double both are at $0/50$; at ViT-Tiny scale on upsampled $64^2$ renders neither model reaches the capability floor, so this cell is uninformative rather than negative.

\section{A scripted reference controller}
\label{app:ceiling}
To calibrate the tier numbers we run a hand-coded controller that reads the simulator state, under exactly the evaluation protocol of \cref{sec:setup}: the same 50 scene seeds per tier, the same scene setup and zero-action warm-up, the same episode budgets (20 or 60 planner steps of five environment steps) and the same success thresholds. On the navigation tier it drives the pusher to the target zone. On the push tiers it orbits the target object at a fixed radius until it is behind the object relative to the target zone, closes in on the contact point, pushes along the object-to-zone direction, and re-plans every environment step from the current positions; it stops once the object is within 12\,px of the zone centre. Results (success rate over 50 seeds): easy $1.00$ / $1.00$, medium $1.00$ / $1.00$, hard $1.00$ / $1.00$, very hard $0.28$ / $1.00$ at 20 / 60 steps. Every push-tier episode moves the target object by more than 5\,px. The very-hard tier is therefore budget-limited even for a controller with state access: its long pushes, often past distractors, rarely fit into 20 planner steps. The same controller solves every tier within 20 steps on both response arms of \cref{app:e1}. The controller is not an optimum, only a demonstration that the scenes are solvable within the budget by a policy with state access, which the learned world models are not.

\section{Displacement statistics of the training distributions}
\label{app:traindist}
\begin{figure}[h]
\centering
\includegraphics[width=0.66\textwidth]{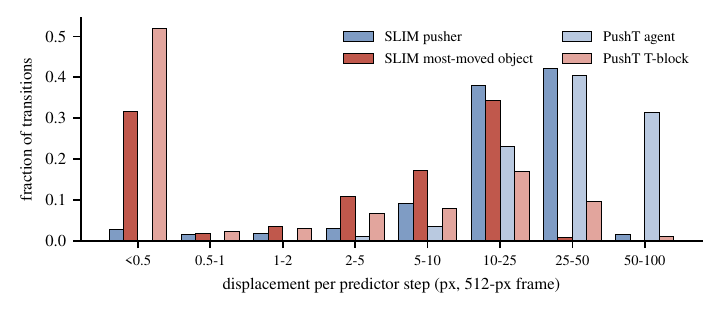}
\caption{Per-predictor-step displacement of pusher/agent and of the most-moved object in the \bench and PushT training data.}
\label{fig:traindist}
\end{figure}

\Cref{sec:bench} characterises the environments under random actions; here we characterise the actual training data. For \bench we replay the dataset generator (same seeds, same scripted primitive policy, 200-step episodes) for 150 episodes without storing frames and record every body position after every environment step; for PushT we read the block and agent positions from the state channel of 1500 random episodes of the standard training release. \Cref{fig:traindist} shows the displacement over one predictor step (frameskip 5) in each simulator's 512-px frame. On \bench ($29{,}400$ transitions) no object moves in $33\%$ of transitions, the most-moved object shifts by under $5$\,px in $48\%$ and under $25$\,px in $99\%$ (mean $7.4$\,px, median $5.7$\,px); the pusher moves $22$\,px in the median. On PushT ($179{,}890$ transitions) the T-block is static in $54\%$ of transitions (median $0.2$\,px) but moves further when it moves (mean $7.8$\,px, 90th percentile $26$\,px), and the agent moves $38$\,px in the median. The fraction of static-object transitions therefore does \emph{not} separate the two benchmarks: a copy-forward prediction of the object is right about as often on PushT as on \bench. What separates them is how much of the frame a moving body changes (the T-block covers far more pixels than a \bench object, so its share of $\Lpred$ is much larger) and how readily the random actions a sampling-based planner draws move it (\cref{sec:bench}: $27$ vs.\ $1.1$\,px root-mean-square displacement over 25 steps). The response-to-actions sweep of \cref{app:e1} tests this account directly by raising the response to random actions at a fixed data policy. Pusher and object positions in the training data are uniform over the arena; push directions are slightly anisotropic (largest deviation of an eight-sector histogram $3.7\sigma$ over 2\,500 episodes) and in single-zone episodes the zone lies in the left half in $82\%$ of cases.

\section{Response-to-actions sweep}
\label{app:e1}
\begin{figure}[h]
\centering
\includegraphics[width=0.8\textwidth]{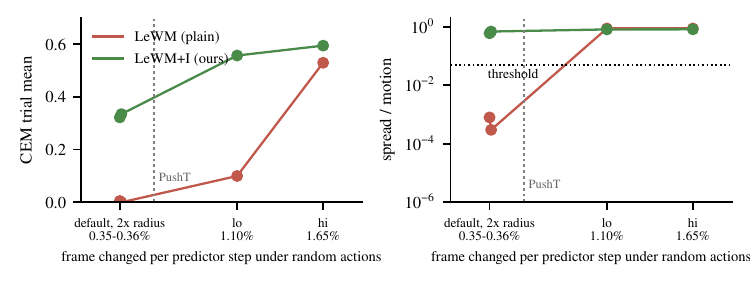}
\caption{Response-to-actions sweep. Left: CEM trial mean; right: action-sensitivity probe (dotted line: usability threshold). Arms ordered by the fraction of the frame that changes per predictor step under native random actions (0.35, 0.36, 1.10, 1.65\%).}
\label{fig:e1}
\end{figure}

We vary how much of the frame responds to sampled actions (\cref{fig:e1,tab:e1}) while holding data volume (10\,000 episodes of 200 steps, base seed 42, the same primitive policy as the main dataset), training (canonical seed 3072, 30 epochs, final checkpoint) and evaluation (the canonical protocol of \cref{sec:exp}, tier generator run under the same physics) fixed. The dose is measured model-free under the environment's native action distribution (under the shared $\sigma{=}0.5$ distribution of \cref{sec:bench} the default scene changes $0.37\%$ and PushT $0.64\%$): the fraction of the $224\times224$ frame whose pixels change per predictor step (16 start states $\times$ 20 sequences of 25 steps; zero actions change $0\%$), and, for reference, the root-mean-square displacement of an object after 25 steps in a single-object scene. Two levers. \emph{Size}: object radius 30\,px ($4\times$ footprint) leaves the changing fraction at $0.36\%$ against $0.35\%$ (object motion $1.1$ vs.\ $0.56$\,px), because larger objects are still rarely hit. Larger radii cannot help further: 45\,px raises object motion only to $1.7$\,px, and from 50\,px the physics breaks (objects overlap at initialisation; zero-action motion $2.7$\,px at 50\,px, $6.9$\,px at 60\,px, exactly zero below). \emph{Response}: at the default 15-px footprint, lighter and less damped objects with a larger and faster pusher, calibrated so that zero-action motion stays zero and the scripted controller of \cref{app:ceiling} solves every tier within 20 steps: \emph{lo} (pusher 28\,px, mass 0.5, damping 0.95, action scale 80; $1.10\%$ of the frame, object motion $6.6$\,px) and \emph{hi} (pusher 40\,px, mass 1, damping 0.95, action scale 80; $1.65\%$, $9.9$\,px). Both the objects and the enlarged pusher contribute to the changing fraction, and we make no claim about which one matters; the claim is about how much of the frame responds. Mass and damping alone do not raise the response much (object motion $3.3$\,px at mass 0.5, damping 0.99): in 24 to 28 of 32 random starts the pusher never touches an object in 25 steps, so contact geometry, not object inertia, is the bottleneck. The response arms are slightly easier for a policy that does not plan: zero actions still score $0.00$, a uniform-random policy $0.045$ (hi) and $0.040$ (lo) trial mean (medium $0.12$ and $0.10$, hard $0.06$ on both; $0.00$ at the default physics), far below the repaired model. At radius 30 the controller solves very hard in $0.30$ of scenes within 20 steps (default $0.28$). At the lo arm plain is already action-sensitive ($0.88$) yet plans poorly ($0.100$), one more latent that clears the gate without being plannable. The \ours cell of the hi arm diverged at the canonical learning rate (validation loss oscillating fourfold from epoch 7) and was retrained at half the rate; all other cells use the canonical setting.

\begin{table}[h]
\centering\small
\resizebox{\textwidth}{!}{%
\begin{tabular}{l ccccc l cccc c c}
\toprule
arm & radius & pusher & mass / damping & scale & frame changed & model & easy & medium & hard & very hard & trial mean & spread/motion \\
\midrule
r15 & 15 & 14 & 8 / 0.85 & 40 & 0.35\% & plain & 0.01 & 0.01 & 0.00 & 0.00 & 0.005 & $8\times10^{-4}$ \\
    &    &    &          &    &        & \ours & 0.96 & 0.20 & 0.13 & 0.00 & 0.323 & 0.60 \\
r30 & 30 & 14 & 8 / 0.85 & 40 & 0.36\% & plain & 0.00 & 0.00 & 0.00 & 0.00 & 0.000 & $3\times10^{-4}$ \\
    &    &    &          &    &        & \ours & 0.99 & 0.22 & 0.12 & 0.01 & 0.335 & 0.69 \\
lo  & 15 & 28 & 0.5 / 0.95 & 80 & 1.10\% & plain & 0.09 & 0.19 & 0.12 & 0.00 & 0.100 & 0.88 \\
    &    &    &          &    &        & \ours & 0.96 & 0.82 & 0.44 & 0.01 & 0.558 & 0.81 \\
hi  & 15 & 40 & 1 / 0.95 & 80 & 1.65\% & plain & 0.83 & 0.80 & 0.40 & 0.09 & 0.530 & 0.88 \\
    &    &    &          &    &        & \ours & 0.98 & 0.93 & 0.40 & 0.07 & 0.595 & 0.82 \\
\bottomrule
\end{tabular}}
\caption{Response-to-actions sweep. One training seed per cell; per-tier values average the two planner horizons; ``frame changed'' is the model-free fraction of the frame that changes per predictor step under native random actions; the r15 rows are the canonical cell of \cref{tab:headline} retrained.}
\label{tab:e1}
\end{table}

\section{Loss-weight sensitivity and the SIGReg weight}
\label{app:alpha}
We retrain the canonical cell at $\alpha_\text{inv}=\alpha_\text{pid}\in\{0.1,0.3\}$ (seed 3072, 30 epochs, canonical protocol) alongside the existing $\alpha=0$ and $\alpha=1$ cells. The repair is graded and monotone in the weight: trial mean $0.000\to0.118\to0.293\to0.368$ for $\alpha=0/0.1/0.3/1$ (per tier easy/medium/hard: $0.41/0.05/0.01$, $1.00/0.10/0.07$, $1.00/0.30/0.17$; very hard $0$ throughout; McNemar $\alpha{=}0.3$ vs.\ $0.1$ and $1$ vs.\ $0.3$ both $p<10^{-5}$). Easy saturates at $\alpha=0.3$; the pushing tiers, including hard where the goal-agnostic baseline scores $0$, keep rising until the weight reaches 1 ($0.01\to0.07\to0.17$). Low weights buy navigation, the full weight buys pushing. At $0.1$, the weight used by DA-LeWM \citep{wang2026dalewm} in a setting where \lewm already plans, the recipe reaches $32\%$ of the canonical result here.

Action sensitivity (environment-path variant of the probe, which gives $0.54$ for the canonical cell against $0.56$ in \cref{tab:probes}) does not track this ordering. $\alpha=0.1$ already lifts spread/motion from $4\times10^{-4}$ to $0.55$, and $\alpha=0.3$ has the highest value of the four cells ($0.80$) at a lower trial mean ($0.293$) than $\alpha=1$ (spread/motion $0.54$, trial mean $0.368$). Since these cells differ only in one scalar, they are the cleanest evidence that the probe is a gate below which every configuration failed rather than a predictor of planning quality above it (\cref{sec:mechanism}).

\paragraph{The SIGReg weight does not cause the collapse.} Retraining plain with $\lambda_\text{sigreg}\in\{0.03,0.27\}$, a third and three times the default $0.09$ (seed 3072, canonical protocol), leaves it at the floor: $1/400$, $0/400$ and $0/400$ successes for $0.03/0.09/0.27$, with spread/motion $0.011$, $4\times10^{-4}$ and $0.002$, all well below the usability gate. The multi-object collapse is therefore not a mistuned regulariser that the IDM gradient happens to compensate for.

\section{Does the latent cost rank actions like the simulator?}
\label{app:spearman}
The probes of \cref{sec:mechanism} are model-internal. This one compares the planner's own cost against the environment. For each of 30 held-out scenes (seeds disjoint from the 50 evaluation seeds) we draw $N=64$ action sequences of $H_a=5$ from the CEM iteration-0 proposal, score each by the latent path cost the planner would use, execute the same sequences in the simulator, and rank-correlate the two. On the easy tier, where the real cost spreads by 145\,px across the proposal sequences so there is something to rank, \ours reaches Spearman $\rho=0.84\pm0.01$, significantly positive on 30 of 30 scenes, while plain is uncorrelated ($-0.02\pm0.05$). On the medium tier the same 25 actions move the object by a median of 2.6\,px, so the diagnostic is signal-starved rather than uninformative: \ours is $+0.12$ against plain's $-0.02$. This is the random-action gap of \cref{sec:bench} seen from the planner's side.

Run across the disentangle cells (easy tier), the diagnostic reproduces \cref{tab:disentangle} with tighter error bars than 400 Bernoulli trials allow: encoder-IDM only $+0.88\pm0.01$ and the full model $+0.84\pm0.01$ (both 30 of 30 scenes significant), against predictor-IDM only $+0.03\pm0.03$ and the shuffled-action null $+0.00\pm0.03$ (at most 4 of 30), both at plain's level. One tension is worth stating: encoder-IDM only \emph{beats} the full recipe on this metric while scoring $7.8$\pp lower planner success on the same canonical checkpoints, so the two measurements disagree about that one ordering. Predictor-IDM may trade a little cost-surface fidelity for a rollout CEM can exploit over 20 steps, but these data do not establish that.

\section{Cross-distribution generalisation}
\label{app:crossdist}
This control separates how little of the frame responds to actions from the presence of distractors. Models trained on a single-object, fixed-appearance \bench variant (one red circle, radius 15\,px) and evaluated on the default multi-object tiers: plain $1/400$; \ours $0.190$ trial mean (easy $0.55$, medium $0.16$, hard $0.05$, very hard $0.00$, pooled over both horizons), about half of the in-distribution $0.368$. Without distractors in training, plain still collapses (spread/motion $6\times10^{-4}$, action-induced object motion $\approx0$), while \ours recovers spread/motion $0.24$, about $400\times$ plain on the same data.

\section{Grounding controls for the language head}
\label{app:layout}
\paragraph{A goal-agnostic policy.} Moving in one fixed direction for $k$ planner steps and then stopping, with direction and $k$ chosen per tier in hindsight, never reads the goal. It solves $0.12$ of easy and $0.20$ of medium scenes and none on hard and very hard (random actions: $0$ on every tier); medium is the tier where the four directions of its axis-aligned template leave the most to a blind policy, so we use the hard tier as evidence of goal-directed pushing.

\paragraph{Where the difficulty lies.} The same MLP design conditioned on the encoded goal frame instead of the sentence, trained on the same $40\,000$ scenes, matches the oracle ($0.363$ vs.\ $0.368$; per tier $0.99/0.29/0.17/0$; McNemar $p=0.80$), so the training pipeline and planner are sound and the difficulty lies in mapping the sentence to a goal latent. A 5-layer MLP language head over the global latent reaches $0.195$ ($0.53/0.16/0.09/0$): lower than the cross-attention head on navigation, higher on medium and hard, and less reliable under the zone swap below (success $0.42$ and $0.48$, two decoy hits, closer to the named zone in 95 of 100 episodes; easy success $0.45$ with the swap vs.\ $0.53$ without).

\paragraph{Zone decoy.} To test whether the head reads the colour in the sentence rather than a typical spatial relation between pusher and target, we swap the target zone with the other zone on the easy tier, so that a zone of the wrong colour occupies the target's position and the named zone sits where the distractor was. A head that reads the colour still succeeds; one that has learned where targets lie goes to the decoy. The visual-goal oracle scores $0.94$ and $0.88$ ($H_a{=}5$ and $10$) on the swapped scenes. Over three seeds the cross-attention head succeeds on $0.84$ of them, exactly as often as without the swap ($252$ vs.\ $253$ of 300), never ends within the threshold of the decoy, and ends closer to the named zone than to the decoy in all 300 episodes (50 scenes, 3 seeds, 2 horizons); per-seed values are in \cref{app:ca}.

\section{Language head: additional results}
\label{app:ca}
\paragraph{Architecture and training.} The cross-attention head has two pre-LN transformer layers (4 heads, hidden 192, FFN 768) over a cls-query initialised from $z_t$, a text-query from a 2-layer refinement of $e_\text{NL}$, and the encoder's 256 patch tokens ($16\times16$ at patch 14) with learned positions; the cls output is $z_\text{goal}$ (1.1M parameters). It is trained for 90 epochs with the mean-squared error to the encoded next-stage latent. Scenes come from the tier generator (seeds from $100\,000$, 2\,500 per tier per $10\,000$ scenes); each of the six frames carries one sentence drawn from 41 stage templates. The MLP head of \cref{tab:nlhead} has 5 layers (hidden 1024, $\sim5$M parameters) over $z_t$ and $e_\text{NL}$ and is trained for 30 epochs.

\paragraph{Stages and sentences.} Training tuples are (latent of the current stage, sentence of the next stage) $\to$ latent of the next stage: the three random pre-contact poses map to the approach stage, approach to midway, midway to goal, and the goal stage to itself. The sentences change with the stage (``Position the pusher behind the red object, aligned with the green zone'', ``Push the red object halfway toward the green zone'', ``The red object is at the green zone''). At test time it receives the final-goal sentence at the start of the episode and again after 10 planner steps, each time from the current frame; \cref{app:staged} describes the staged protocol of \cref{sec:staged}.

\paragraph{Paraphrase pools.} The train pool has 107 templates, a superset of the 41 training templates with 66 rephrasings in the same vocabulary style; the test pool has 31 held-out templates whose object and zone nouns never occur in training, with verbs overlapping partly, mostly on the navigation templates. Per tier on seed 42 (easy/medium/hard): canonical $0.88/0.05/0.05$, train pool $0.84/0.04/0.02$, test pool $0.78/0.01/0.03$; very hard is $0$ throughout.

\paragraph{Zone decoy on three seeds.} Easy tier, $H_a{=}5/10$, 50 seeds each: success $0.92/0.86$, $0.84/0.68$, $0.92/0.82$ for head seeds 42, 7, 11 (oracle $0.94/0.88$); decoy hits $0$ in all 300 episodes; ended closer to the named zone in $100$ of 100 for every seed. Without the swap the same heads score $0.88$, $0.78$ and $0.87$ on easy.

\section{Staged instructions: protocol and additional results}
\label{app:staged}
\paragraph{Protocol.} The stage sentences are the canonical templates of the head's training data: for pushing ``Position the pusher behind the \{object\} object, aligned with the \{zone\} zone.'', ``Push the \{object\} object halfway toward the \{zone\} zone.'' and ``The \{object\} object is at the \{zone\} zone.'', for navigation ``Move the pusher halfway toward the \{zone\} location.'' and ``The pusher has reached the \{zone\} location.'', filled with the colours of the target object and zone. Query $k$, at planner step $7k$, uses stage $\min(k,\text{last})$: pushing receives the approach, midway and goal sentences at steps 0, 7 and 14, navigation the approach sentence at step 0 and the goal sentence at steps 7 and 14. Each query is made from the current frame, and the planner uses the latest output. The stage frames of the oracle are rendered from the scene configuration with the target object and the pusher placed as in the head's training frames: pusher at contact distance ($29$\,px) behind the object on the line from object to zone (approach), object at the midpoint between its start and the zone with the pusher behind it (midway), object at the zone with the pusher behind it (goal, identical to the visual oracle's goal frame); for navigation the midpoint between pusher start and target, then the target. All other objects stay at their start positions.

\paragraph{Latent switch.} The switch rule is evaluated at every planner step; the head is queried only when a stage is entered, from the frame at that step, and its output is kept as the target until the switch. With $d_t$ the mean squared distance between the current latent and the stage target and $d_0$ its value at entry, the rule advances when $d_t\le\rho\,d_0$ or when $d_t$ has not reached a new minimum for $P$ consecutive steps. We searched $\rho\in\{0.1,0.25\}$ and $P\in\{2,3,5\}$ on 20 tuning scenes per tier (seeds 1000--1019, disjoint from the evaluation seeds; easy, medium and hard; $H_a{=}5$; head seed 42), where the six settings reach 33--36 of 60 successes and the fixed schedule 33, and use $\rho=0.1$, $P=3$ (tied with $P=5$; we took the middle of the grid). On the evaluation scenes the rule enters the midway stage after about 5 planner steps and the goal stage after about 11 on medium and hard (fixed schedule: 7 and 14), and the goal stage after about 3 steps on navigation; at least $98\%$ of the pushing episodes reach the goal stage for every head seed.

\paragraph{Simulator-judged switch.} This reference advances from approach once the pusher is within the success threshold ($30$\,px) of the approach pose or the object has closed $15\%$ of its distance to the zone, from midway once the object has closed half of that distance, and on navigation once the pusher has closed half of its distance to the target; stage targets are computed on entry as for the latent switch. On head seed 42, $91\%$ of the medium and $80\%$ of the hard episodes end in the midway stage and never receive the goal sentence; 12 of the 13 medium and hard episodes that do reach the goal stage succeed.

\paragraph{Stage mapping.} \Cref{tab:stagemap} queries the head as the staged evaluation does, but from rendered stage frames: the approach sentence from the start frame, the midway sentence from the approach frame and the goal sentence from the midway frame of each evaluation scene. A query counts as correct if its output is nearer to the encoded frame of the intended stage than to the other stage frames. On navigation the approach query, whose target is the midpoint, is the least precise. From the start frame the goal sentence moves the latent most of the way toward the encoded goal frame (a remaining distance of $0.10$--$0.12$ on medium and hard, where the start frame lies at $0.36$--$0.39$), but its error is $2.3$ to $3.5$ times that of the same query from the midway frame ($0.03$--$0.05$, where the midway frame lies at $0.08$--$0.09$).

\begin{table}[h]
\centering\small
\setlength{\tabcolsep}{4pt}
\begin{tabular}{l cc ccc ccc ccc}
\toprule
& \multicolumn{2}{c}{easy} & \multicolumn{3}{c}{medium} & \multicolumn{3}{c}{hard} & \multicolumn{3}{c}{very hard} \\
\cmidrule(lr){2-3}\cmidrule(lr){4-6}\cmidrule(lr){7-9}\cmidrule(lr){10-12}
head seed & appr. & goal & appr. & mid. & goal & appr. & mid. & goal & appr. & mid. & goal \\
\midrule
42 & 0.74 & 0.98 & 0.96 & 1.00 & 0.94 & 0.94 & 1.00 & 0.94 & 1.00 & 1.00 & 0.98 \\
7  & 0.66 & 1.00 & 0.90 & 0.98 & 0.92 & 0.76 & 1.00 & 0.92 & 1.00 & 1.00 & 0.98 \\
11 & 0.64 & 0.98 & 1.00 & 1.00 & 0.96 & 1.00 & 1.00 & 0.94 & 1.00 & 1.00 & 0.98 \\
\bottomrule
\end{tabular}
\caption{\textbf{Stage mapping}: fraction of the 50 evaluation scenes per tier in which the head's output for the next-stage sentence, queried from the rendered frame of the previous stage, lies nearest to the encoded frame of the intended stage.}
\label{tab:stagemap}
\end{table}

\paragraph{Per-seed values.} \Cref{tab:stagedseeds} lists the language rows of \cref{tab:staged} per head seed. Stage sentences against the goal sentence: $51/13$, $46/19$ and $69/8$ discordant trials (head seeds 42, 7, 11); against the goal-frame oracle over all tiers: $p=0.64$, $2\times10^{-4}$ and $0.067$, where seed 7 falls short on navigation (easy $0.72$ vs.\ $0.98$); on the pushing tiers alone no seed is below the oracle ($p=0.23$ and $0.45$ for seeds 42 and 7) and seed 11 is above it ($p=0.003$). Latent switch against the fixed schedule: $p=0.29$, $0.31$ and $0.06$. Zone decoy with stage sentences, same campaign ($H_a{=}5/10$): success $0.94/0.82$, $0.78/0.80$ and $0.90/0.88$; with the goal sentence in the same run $0.88/0.90$, $0.90/0.82$ and $0.86/0.92$; no episode ends at the decoy, and every episode ends closer to the named zone.

\begin{table}[h]
\centering\small
\setlength{\tabcolsep}{4pt}
\resizebox{\textwidth}{!}{%
\begin{tabular}{l ccc}
\toprule
trial mean (easy / medium / hard / very hard) & head seed 42 & head seed 7 & head seed 11 \\
\midrule
goal sentence & $0.245$ (0.88/0.04/0.06/0) & $0.205$ (0.76/0.04/0.02/0) & $0.238$ (0.87/0.04/0.04/0) \\
stage sentences, every 7 steps & $0.340$ (0.83/0.27/0.25/0.01) & $0.273$ (0.72/0.17/0.19/0.01) & $0.390$ (0.91/0.32/0.31/0.02) \\
stage sentences, latent switch & $0.363$ (0.90/0.33/0.21/0.01) & $0.293$ (0.76/0.21/0.20/0) & $0.350$ (0.80/0.32/0.28/0) \\
paraphrases, train pool & $0.353$ (0.87/0.30/0.22/0.02) & $0.245$ (0.68/0.20/0.10/0) & $0.318$ (0.80/0.23/0.22/0.02) \\
paraphrases, unseen nouns & $0.238$ (0.74/0.15/0.06/0) & $0.215$ (0.69/0.09/0.08/0) & $0.218$ (0.80/0.04/0.03/0) \\
\bottomrule
\end{tabular}}
\caption{\textbf{Staged instructions per head seed} (400 trials each).}
\label{tab:stagedseeds}
\end{table}

\end{document}